\pdfoutput=1

\documentclass[11pt]{article}

\usepackage[]{acl}

\usepackage{times}
\usepackage{latexsym}
\usepackage{todonotes}
\usepackage{hyperref}

\usepackage{titlesec}

\usepackage[T1]{fontenc}

\usepackage[utf8]{inputenc}

\usepackage{microtype}
\usepackage{inconsolata}
\usepackage{graphicx}
\usepackage{placeins}
\usepackage{xurl}
\usepackage[bottom]{footmisc}
\usepackage[utf8]{inputenc}
\usepackage{lipsum,caption,graphicx}
\usepackage{multirow}
\usepackage{booktabs}
\usepackage{afterpage}
\usepackage{adjustbox}
\usepackage{tabularx}
\usepackage{subfig}
\usepackage{xr}
\usepackage{xspace}
\usepackage{tablefootnote}
\usepackage{caption}
\usepackage{amsmath}
\usepackage{hyperref}
\usepackage{adjustbox}
\usepackage{color,soul}
\usepackage{amsmath}
\usepackage{xcolor}
\usepackage{float}
\usepackage[bottom]{footmisc}

\newcommand{\mycomment}[1]{}
\newcommand{\floor}[1]{\lfloor #1 \rfloor}
\title{A Template Is All You Meme}

\author{Luke Bates,$^1$ Peter Ebert Christensen,$^{2,3}$ Preslav Nakov,$^4$ and Iryna Gurevych$^1$ \\
  $^1$Ubiquitous Knowledge Processing Lab (UKP Lab)\protect\\ Department of Computer Science and Hessian Center for AI (hessian.AI)\protect\\ Technical University of Darmstadt \protect\\ 
  $^2$Department of Computer Science, University of Copenhagen, $^3$Pioneer Centre for AI\protect\\
  $^4$Mohamed bin Zayed University of Artificial Intelligence\protect\\
  \url{https://www.ukp.tu-darmstadt.de/}\protect\\ 
}

\begin{document}
\maketitle
\begin{abstract}
Templatic memes, characterized by a semantic structure adaptable to the creator's intent, represent a significant yet underexplored area within meme processing literature. With the goal of establishing a new direction for computational meme analysis, here we create a knowledge base composed of more than 5,200 meme templates, information about them, and 54,000 examples of template instances (templatic memes). To investigate the semantic signal of meme templates, we show that we can match memes in datasets to base templates contained in our knowledge base with a distance-based lookup. To demonstrate the power of meme templates, we create TSplit, a method to reorganize datasets, where a template or templatic instance can only appear in either the training or test split. Our re-split datasets enhance general meme knowledge and improve sample efficiency, leading to more robust models. Our examination of meme templates results in state-of-the-art performance for every dataset we consider, paving the way for analysis grounded in \emph{templateness}.\footnote{Our code is available at \url{https://github.com/UKPLab/naacl2025-a-template-is-all-you-meme}.}

\end{abstract}
\textbf{WARNING: For demonstration purposes, we show memes that some may find offensive.} 

\section{Introduction}
Memes are a form of communication capable of succinctly conveying complicated messages. They have been defined as a unit of cultural transmission or of imitation and replication \cite{r.dawkins1976the-selfish-gen}, but
all memes possess the trait of referencing a cultural moment shared by a group of people. Despite their basis in Internet culture, they exhibit sociolinguistic traits of in-group communication \cite{memeling, Holm_2021_sociolinguistic_memes}. A meme's meaning can therefore be opaque to those who do not belong to the in-group.

\begin{figure}[t]%
    \centering
    \includegraphics[width=.45\textwidth]{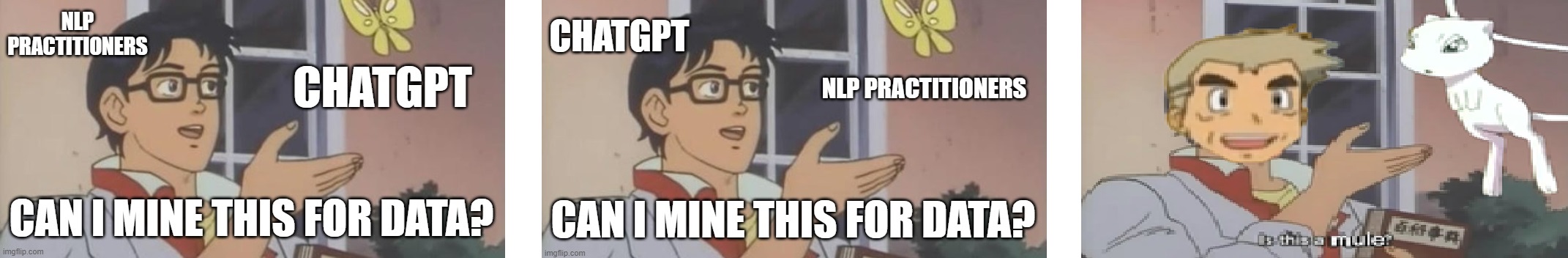}
    \caption{The meaning of templatic memes is customizable via overlaid text or image(s), but remains grounded in the context of the template. The first panel suggests that the NLP community thinks it can use ChatGPT to generate data, while the second one suggests that ChatGPT can exploit the NLP community for data. The third one uses overlaid images to reference Pokemon.}
    \label{pigeon1}
    
\end{figure}

\begin{figure*}[t]%
    \centering
    \includegraphics[width=.85\textwidth]{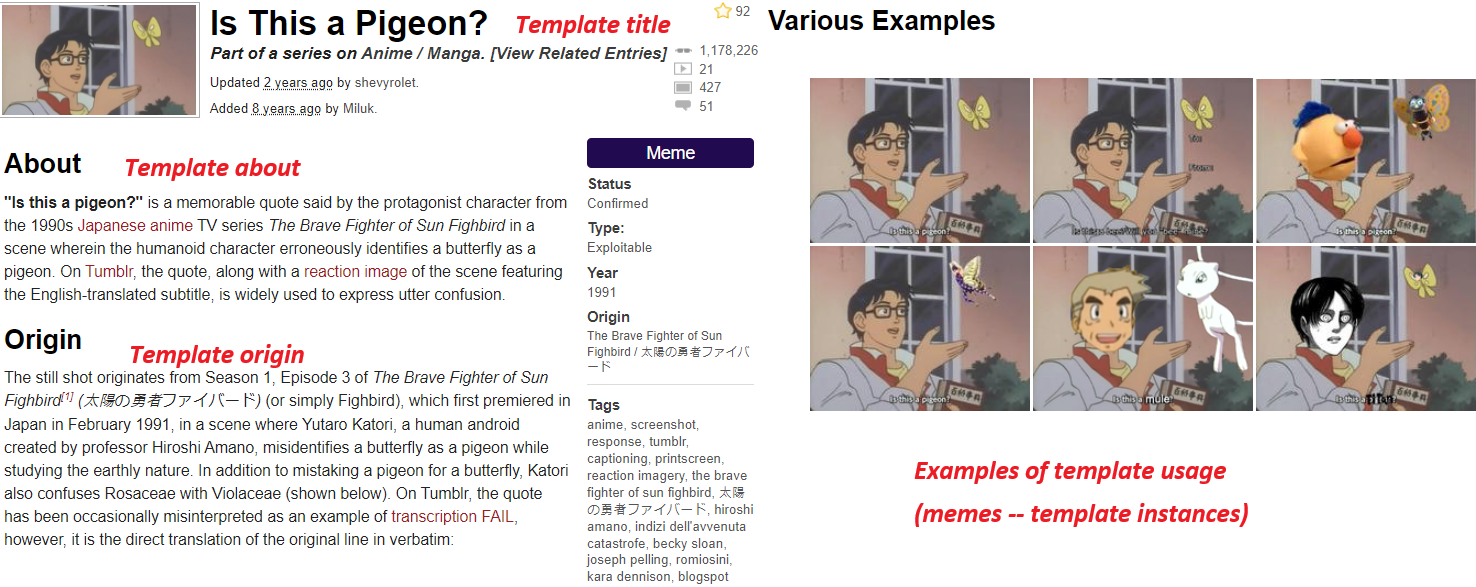}
    \caption{Example entry from KYM where we have labeled relevant data fields in red.}
    \label{pigeon2}
\end{figure*}
Meme templates are common patterns or elements, such as text or images, that are used to create novel memes. They can be difficult to parse because they can be combined in different ways and each one has its own unique meaning, the specific semantics of which is customizable by the person posting the meme (the \emph{poster}). The template and its message can be referenced by an image, but may not be directly related to that image. If the viewer is not familiar with the template in question, they may not understand the meme's meaning. For example, in Figure \ref{pigeon1}, we see three instances of the popular \emph{Is This a Pigeon?}\footnote{\url{https://knowyourmeme.com/memes/is-this-a-pigeon}} template. This template conveys the idea that the subject of the person is misinterpreting the object of the butterfly due to his own worldview or limited knowledge. The meaning can be tuned by the poster using overlaid text or images. Importantly, such altered images are considered instances of the same template. To interpret memes, one must recognize the entities in the meme and the template the meme uses, if any. 

It is important to distinguish between templatic memes and other meme types. Templatic memes reference a meme template, which is a commonly reused material (e.g.,~text, images, audio), to create a novel instance that is still grounded in the meme template's semantics. Non-templatic memes can be (visual) puns, jokes, or emphasis that might not directly reference a meme template and can be understood without knowledge about meme templates or even memes (see Section \ref{notemplates}).

\emph{Know Your Meme} (KYM),\footnote{\url{https://knowyourmeme.com/}} the Internet Meme Database, is a valuable resource for information related to memes, and especially, to templatic memes. Even people familiar with memes may not be aware of a specific template and can look it up on KYM. The meme entries provide the base template and information about it, such as its meaning, origin, examples, etc. By reviewing entries of unfamiliar templates, users learn how to interpret and use the template themselves to create novel instances for their specific communication needs.

Memes are of interest to the machine learning community \citep{Aggarwal-2023} because they are difficult, but can still be formulated as classification or generation tasks \cite{peirson2018dank}. They are also used to spread misinformation and hate speech \cite{pramanick-etal-2021-detecting}. Memes usually express concepts humorously, and humor has been shown to increase the persuasiveness of an idea \cite{10.1093/hcr/hqy005}. Thus, developing systems that can understand memes is important to prevent the spread of harmful content. However, AI researchers and datasets treat memes as static images with text \cite{visual-memes, qu2022disinfomeme} despite memes having their own unique characteristics, such as being templatic or not.

Here, we analyze memes as memes, rather than as images, text, or a combination of both. To support this, we introduce the Know Your Meme Knowledge Base (KYMKB), a database of meme templates and associated information sourced from KYM. We hypothesize that knowledge of meme templates provides valuable context for understanding memes. Using a distance-based lookup, we match templatic instances to base templates in the KYMKB, enabling retrieval of meme-specific context. This context is useful for prompting large language models (LLMs) or multimodal Vision-LMs in a retrieval-augmented generation (RAG) framework, as demonstrated in Section \ref{prompting}.

To investigate the effect meme templates have on modelling, we create a method for (re)splitting meme datasets such that a template and its instances can only appear in either the training or test split, which we call the Template-Aware Splitter (TSplit). TSplit (re)organizes dataset entries based on their distance in embedding space from the KYMKB. We find that controlling for \emph{template awareness} makes TSplit more sample-efficient than standard fine-tuning or random sampling. We also show that TSplit has strong regularization effects in scenarios where models are likely to overfit.

Our analysis of meme templates results in state-of-the-art (SOTA) performance for all datasets we consider, establishing a robust foundation for future research grounded in \emph{templateness}.
Our contributions are as follows:
\begin{enumerate}
    \item We suggest a new direction for future meme analysis grounded in templates. 
    \item We construct a knowledge base of meme templates and information about them.
    \item We show meme templates are useful for a number of tasks and can have marked effects in fine-tuning experiments.
\end{enumerate}

\section{Related Work}

There has been a lot of work on analyzing memes in various task formulations.
\subparagraph{Memes as harmful content} This includes MultiOFF \cite{suryawanshi-etal-2020-multimodal}, a dataset of offensive memes related to the 2016 US presidential election. 
The MAMI dataset \cite{fersini-etal-2022-semeval} is composed of two datasets from SemEval-2022: in subtask A, the goal is to identify misogyny in memes, while in subtask B, it is to determine different types of misogyny expressed by a meme. \citet{lin-etal-2023-beneath} recognized that the surface-level text and the image of memes are insufficient and employed large language model (LLM) knowledge distillation to classify dangerous memes.  

\subparagraph{Memes as a form of language} \citet{dimitrov-etal-2021-semeval} pointed out that memes can be persuasive by exploiting more than 20 different propaganda techniques. FigMemes \cite{liu-etal-2022-figmemes} scraped images from a politically incorrect and infamously toxic board on 4chan, /pol/,\footnote{\url{https://boards.4chan.org/pol/}} and labeled over five thousand memes with six different types of figurative language used in the meme, recognizing that memes are capable of expressing abstract and complicated messages. \citet{mishra2023memotion} released Memotion 3, which is composed of memes in Hindi and English, labeled for sentiment, emotion detection, and emotion intensity. \citet{hwang-shwartz-2023-memecap} released a dataset of meme explanations to aid in resolving metaphors in memes. \citet{zhou-etal-2024-social} showed that memes and meme templates can be clustered by their semantic function.

\subparagraph{Context for memes} All the above work fine-tuned multimodal PLMs or prompted LLMs on their respective datasets, but did not use additional context in order to increase meme understanding. This is a trend in meme-related ML research. One exception is MEMEX \cite{sharma-etal-2023-memex}. They use Wikipedia and Quora to assemble explanations to ask if an explanation document is relevant for a meme, formulating a novel task and multimodal model. Notably, this work uses meme-external information (Wikipedia/Quora), but not meme knowledge, e.g.,~information about the template used by the meme. We emphasize that the context they inject is common knowledge or knowledge about named entities, not knowledge about memes.

\subparagraph{General meme resources} \citet{DBLP:conf/esws/TommasiniIW23} developed a knowledge graph of memes by scraping and querying different sources of information, such as KYM, to connect memes to the information they reference. However, they do not leverage the graph in a downstream task, nor is it clear how their graph could be applied to meme analysis due to a lack of demonstrations. 

\subparagraph{The current work} Our work focuses on the meme-specific phenomenon of meme templates. To this end, we construct a knowledge base from KYM to ground our analysis in a meme-specific context. The KYMKB is not a dataset and is not labeled for a specific task, but contains information about templatic memes, such as their title, meaning, and origin. We show that we can use CLIP \cite{pmlr-v139-radford21a} and distanced-based lookup algorithms ``off the shelf'' and match templates to memes in existing datasets. To investigate the effect meme templates have on meme processing, we develop TSplit for  (re)organizing datasets based on \emph{templateness}, the Euclidean distance of meme-vector representations from the KYMKB. This has significant effects in fine-tuning experiments, establishing SOTA performance in multiple setups.

\begin{figure*}[t]%
    \centering
    \includegraphics[width=.85\textwidth]{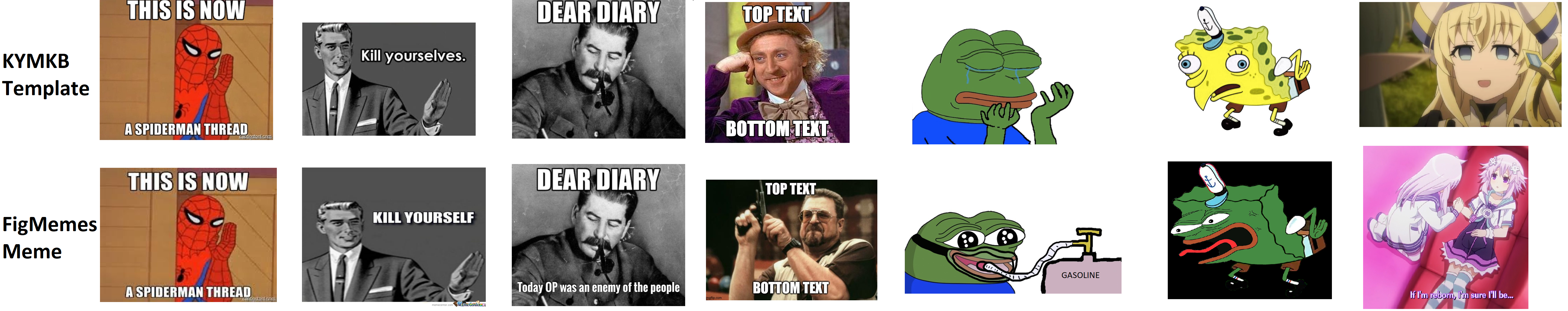}
    \caption{KYMKB templates (first row) vs. their nearest neighbor in the FigMemes dataset (second row).}
    \label{retrival}
\end{figure*}

\section{The Know Your Meme Knowledge Base}
\emph{Know Your Meme}, the ``Internet Meme Database'', can be thought of as the Wikipedia for memes. Users create web pages of a meme template and document information about it, e.g.,~its meaning, and add examples of its usage (see Figure~\ref{pigeon2}). The community reviews and approves entries, updating them as the template's usage evolves. 

Template instances are important for meme understanding. In Figure~\ref{pigeon2}, we see that the template can be altered via overlaid text and images to tune it for a specific communication goal. Existing approaches rely on OCR to extract the text and/or the named entities \cite{kougia2023memegraphs}, but this would not work in many cases, e.g., if the entities are images referencing a popular YouTube video.\footnote{\url{https://www.youtube.com/watch?v=sXOdn6vLCuU&t=8s}}

KYM is a valuable resource that has been under-utilized by the AI community. To address this, we create the KYMKB, a collection of meme templates, examples, and information about the meme's usage. To ensure the quality of the entries, we crawl templates from KYM that are approved by the community, scraping 5,220 base templates and 49,531 examples (see Section \ref{scrape}).


Memes may deviate from their template-based origin, and the KYMKB accounts for this. Consider our running example of \emph{Is This a Pigeon?}. This template was popularized in 2011, but it then had a resurgence in April of 2018. By June 2018, a female version of the template had emerged, which was interpreted by some users as an example of gender transitioning. Such usage and evolution is documented by KYM users in the form of both text and images.  Popular template instances that differ from their origin often become their own template, such as \emph{Pepe the Frog}\footnote{\url{https://knowyourmeme.com/memes/pepe-the-frog}} vs. \emph{Feels Bad Man/Sad Frog}.\footnote{\url{https://knowyourmeme.com/memes/feels-bad-man-sad-frog}} The former is a template originally used in a manner similar to emoticons, while the latter is a popular instance of Pepe that became its own template expressing sorrow or disappointment. By collecting examples, user-curated information, and distinct but related templates, the KYMKB is organized for the dynamic nature of memes.



\section{Template-Meme Analysis}
\label{eda}
To investigate the power of templates, we conducted exploratory data analysis on four meme datasets, MultiOff, Memotion 3, FigMemes, and MAMI. Specifically, we encoded the KYMKB templates, fit a nearest neighbor lookup, as this is an intuitive and commonly used vector-similarity measure \citep{sklearn-api}, query the 500 closest neighbors in the datasets, and manually inspect the similarities between template-meme pairs. Henceforth, we use CLIP as our encoder as it is a commonly-used PLM for vision and language learning problems and memes \citep{pramanick-etal-2021-momenta-multimodal}, but the encoding function is ultimately arbitrary and we refer to it as $f$. In our analysis, we group memes into four categories, detailed below:

\begin{enumerate}
    \item \textbf{Base} - the meme is the base template or cropped/distorted version of it.
    \item \textbf{Instance} - the meme is a templatic instance customized with text or overlaid images.
    \item \textbf{Relevant} - the meme employs more than one template or an obvious reference to a template.
    \item \textbf{Irrelevant} - the meme/image does not appear related to its template pair.
\end{enumerate}

\begin{table}[h]
\centering
    \begin{adjustbox}{width=.45\textwidth}
    \begin{tabular}{lccccc}
    \toprule
    \textbf{Dataset} & \textbf{Base} & \textbf{Instance} & \textbf{Relevant} & \textbf{Irrelevant}\\
     \midrule
     MultiOff & 10.4 & 30 & 21.3 & 38.3    \\
     Memotion 3 & 13.4 & 48.2 & 11.6 & 26.8   \\
     FigMemes & 39.2 & 15.2 & 28.8 & 16.8 \\
     MAMI & 5.7 & 24.1 & 20.3 & 49.9 \\ 
     \midrule
    \end{tabular}
    \end{adjustbox}
    \caption{Percentages of memes grouped by visual similarities to their nearest neighbor in the KYMKB.}
    \label{tab_eda}
\end{table}
\noindent

Figure \ref{retrival} shows a sample of our template-meme pairs and Table \ref{tab_eda} summarizes our groupings in percentages. In many cases, the memes we find are our base templates or distorted versions of them, such as the first two columns in the figure. We observed many instances of templates tuned by the poster, such as the third column. To validate our grouping, we sampled 50 template-meme pairs and asked two of our colleagues to annotate them based on our four categories. For the \emph{Base}, \emph{Instance}, \emph{Relevant}, and \emph{Irrelevant} groupings, we computed a Cohen's Kappa of 1.0, 0.45, -0.11, and 0.80 respectively, with the average being 0.54, moderate agreement. It is reasonable that our annotators could be confused by the \emph{Instance} and \emph{Relevant} grouping as this distinction is subjective, however, we note the perfect agreement and near perfect agreement we captured for the \emph{Base} and \emph{Irrelevant} grouping.

\begin{table*}[t]
\centering
    \begin{adjustbox}{width=\textwidth}
    \begin{tabular}{lccccccc}
    \toprule
    \textbf{Dataset} & \textbf{Original Training Size} & \textbf{TSplit Training Size} & \textbf{Original Validation Size} & \textbf{TSplit Validation Size} & \textbf{Original Test Size} & \textbf{TSplit Dummy Test Size}\\
     \midrule
     MultiOff & 445 & 381 & 149 & 96  & 149 & 117   \\
     Memotion 3 & 5600 & 4674  & 1400 & 1169  & 1500 & 1157   \\
     FigMemes & 3084 & 2333 & 515 & 260 & 1542 & 1006  \\
     MAMI & 8000 & 7353 & 2000 &  1839  & 1000 & 808 \\ 
    \midrule
    \end{tabular}
    \end{adjustbox}
   \caption{Original dataset split sizes and example TSplit dataset splits derived from training and validation data. The dummy test data was discarded. \texttt{ViT-L/14@336px} was used as the encoder. See Table \ref{tab_tsplitsizes_encoder} for other encoders.}
    \label{tab_tsplitsizes}
\end{table*}
\noindent
We are able to match templates to relevant instances despite different appearances. For example, the KYMKB includes \emph{Pepe the Frog}, a template with many different versions, which is also a symbol of the alt-right movement \cite{pepe}. When we query FigMemes, we capture an instance of a happy Pepe inhaling gasoline, communicating the idea that only death can bring the poster happiness. Going a step further, we see two templatic concepts merging into a single meme. \emph{Mocking SpongeBob}\footnote{\url{https://knowyourmeme.com/memes/mocking-spongebob}} is a popular template, which is used to express contempt. The nearest neighbor to this template in FigMemes is an instance where SpongeBob has been amalgamated with the angry Pepe. Querying the KYMKB retrieves enough information from the \textit{about} sections to interpret this meme as the alt-right angrily expressing derision, consistent with /pol/ \cite{kek}, the domain from which FigMemes was created.

KYMKB may match a meme or image that is not a template instance. The $7^{th}$ column shows the template of \emph{You Get Used To It}\footnote{\url{https://knowyourmeme.com/memes/you-get-used-to-it}} matched to a picture that appears to be a still from another anime.\footnote{\url{https://en.wikipedia.org/wiki/Hyperdimension_Neptunia}} The FigMemes image is not an instance of the aforementioned template, but this is subjective as it is not possible to know every template nor does the KYMKB encompass all meme knowledge. We note that for MAMI this especially seems to be the case, where for nearly half of the examples we considered, our template-meme pairs did not appear to be related (see Table \ref{tab_eda}). Not all memes reference a template and meme datasets contain non-meme images (see Figure \ref{non_templatic}).

\section{Template-Aware Splitter}
\label{tsplit}

Given that we are able to successfully match memes to templates in the KYMKB, we hypothesize that many meme datasets are composed of nothing more than templatic instances, which may already be contained within our knowledge base. This would then create a strong signal we could leverage in a downstream task.

To investigate this, we demonstrate another use case and develop a tool from the KYMKB called the Template-Aware Splitter (TSplit), which quantifies \textit{templateness} by comparing base templates to their examples (see Figure \ref{pigeon2}). TSplit embeds a template ($ref = f(KYMKB_{[i]})$) and its examples ($examp = f(KYMKB_{[i][j]})$), then computes the Euclidean distance between them ($dists_{[i]} = dist(ref, examp_{[j]})$). These distances are used to compute a threshold value, such as the median distance ($threshold_{[i]} = median(dists_{[i]})$). We explore four threshold methods: maximum, median, mean, and 25$^{th}$ percentile distances. Using the maximum provides a lenient definition of a template instance, allowing greater visual variance (see Figure~\ref{retrival}), while the 25$^{th}$ percentile offers a stricter interpretation. TSplit encodes a meme, matches it to the closest template in the KYMKB, and compares the distance to the template’s threshold. If the distance exceeds the threshold, the meme is assigned a unique identifier (UI); otherwise, it is recorded as an instance of the template. While our focus is on templatic memes, TSplit also handles non-templatic memes via UIs.

We use TSplit to reorganize datasets and assign memes to a template or declare them non-templatic. TSplit samples our templates and UIs without replacement, such that a template or UI cannot appear in both the training and the test data. Our goal is to decouple the data distribution from template robustness, inspired by work done in QA, NLI, and bias detection on adversarial dataset creation and modelling \cite{jia-liang-2017-adversarial, gururangan-etal-2018-annotation, baly-etal-2020-detect}. Note that TSplit is task agnostic (see Section \ref{tsplit_deets}). 

To examine the effect TSplit has on dataset sampling, we tested various versions of TSplit on six meme classification datasets: FigMemes with seven labels of different types of figurative language, MultiOff a binary dataset where the labels can be offensive or non-offensive, Memotion 3, where Task A has three labels for sentiment analysis and Task B has four for different types of emotion, and MAMI with Task A being  binary misogyny detection and Task B having four labels with different types of misogyny being expressed (see Tables~\ref{datasum} and ~\ref{modelsum}). We choose these datasets because they were made to address the issue of harmful meme detection and to frame our analysis of meme templates in the light of this important goal. Each dataset used Google Vision to provide OCR text overlaid on memes.\footnote{\url{https://cloud.google.com/use-cases/ocr}}

To avoid overlapping templates and unique identifiers between dataset splits, we construct an array of distinct objects, meaning detected templates or unique identifiers. We randomly shuffle the array and then create a test split index. Everything that appears before the index is an object that can appear in the training data and everything after can appear in the test data. We use the following formula to create this index: $\floor{(\frac{t_{size}}{d_{size}})* o_{size}}$, where $t_{size}$ is the number of test examples in the original dataset, $d_{size}$ is the total size of the dataset, and $o_{size}$ is the number of distinct detected objects (templates or unique identifiers).
We maintain the original dataset ratios (for example training 60\%, validation 20\%, and test 20\%), but because we sample the resplit datasets based on detected templates, it is difficult to maintain the exact numbers.

Finally, it is possible that a template in the KYMKB does not have any examples. In such cases, we use a global threshold value calculated by taking the maximum, median, mean, or $25^{th}$ percentile value of all template thresholds.

\begin{table*} [t]
\centering
    \begin{adjustbox}{width=.85\textwidth}
    \begin{tabular}{lccccccc}
    \toprule
    \textbf{Split} & \textbf{MultiOff} & \textbf{Memotion 3 (A)} &\textbf{Memotion 3 (B)} & \textbf{FigMemes} & \textbf{MAMI (A)} & \textbf{MAMI (B)} & \textbf{Overall} \\
    \midrule
    & & & \textbf{Encoder: \texttt{ViT-L/14@336px}}& & & & \\
    \midrule
Original$_{ViT-L/14@336px}$ & 63.64$_{2.31}$ & 26.38$_{1.57}$ & 81.7$_{2.33}$ & 47.79$_{1.43}$ & \textbf{71.82}$_{4.05}$ & 56.6$_{1.56}$ & $57.99_{17.76}$ \\
TSplit$_{max}$ & \textbf{66.16}$_{4.11}$ & 26.68$_{1.29}$ & 83.98$_{1.76}$ & 47.76$_{0.8}$ & 68.72$_{2.21}$ & \textbf{58.69}$_{1.56}$ & $\textbf{58.66}_{17.98}$\\
TSplit$_{median}$ & 64.54$_{3.0}$ & \underline{28.35}$_{1.73}$ & 82.93$_{3.15}$ & \textbf{48.85}$_{1.31}$ & 68.63$_{1.25}$ & 58.19$_{2.13}$ & $58.58_{17.02}$\\
TSplit$_{mean}$ & 64.21$_{3.87}$ & 28.14$_{1.21}$ & 82.57$_{1.49}$ & 47.43$_{1.52}$ & 68.62$_{2.42}$ & 57.41$_{1.88}$ & $58.06_{17.12}$ \\
TSplit$_{percentile}$ & 62.04$_{3.36}$ & 26.79$_{2.08}$ & 83.24$_{3.39}$ & 44.63$_{1.96}$ & 68.87$_{0.99}$ & 58.12$_{0.98}$ & $57.28_{17.89}$\\
Baseline$_{ViT-L/14@336px}$ & 65.03$_{2.66}$ & 26.36$_{1.05}$ & \underline{84.31}$_{1.5}$ & 45.21$_{2.52}$ & 70.68$_{2.86}$ & 57.73$_{1.05}$ & 58.22$_{18.56}$\\
\midrule 
& & & \textbf{Encoder: \texttt{ViT-B/32}}& & & & \\
\midrule
Original$_{ViT-B/32}$ & 60.34$_{2.21}$ & \textbf{28.46}$_{1.67}$ & 82.65$_{1.04}$ & 38.26$_{1.59}$ & \underline{68.17}$_{1.05}$ & 54.0$_{0.82}$ & \underline{55.31}$_{18.03}$ \\
TSplit$_{max}$ & \underline{60.4}$_{2.17}$ & 26.02$_{0.72}$ & 82.0$_{2.44}$ & 37.44$_{1.3}$ & 67.59$_{1.54}$ & \underline{55.28}$_{1.37}$ & $54.79_{18.55}$ \\
TSplit$_{median}$ & 58.61$_{3.08}$ & 26.66$_{0.72}$ & 82.24$_{1.89}$ & 37.74$_{2.21}$ & 65.79$_{2.08}$ & $54.05_{1.8}$ & $54.18_{18.12}$ \\
TSplit$_{mean}$ & 58.24$_{2.17}$ & 25.98$_{0.94}$ & \textbf{84.35}$_{2.22}$ & 38.98$_{0.83}$ & 65.61$_{2.75}$ & 54.61$_{1.85}$ & $54.63_{18.63}$ \\
TSplit$_{percentile}$ & 60.09$_{2.15}$ & 25.83$_{0.84}$ & 81.61$_{2.05}$ & \underline{39.22}$_{2.15}$ & 67.16$_{2.79}$ & 54.49$_{1.81}$ & $54.73_{18.17}$\\
Baseline$_{ViT-B/32}$  & 60.07$_{3.29}$ & 27.78$_{1.38}$ & 83.31$_{0.48}$ & 39.47$_{1.45}$ & 67.11$_{1.77}$& 53.5$_{1.38}$ & 55.21$_{18.06}$\\
\midrule 
& & & \textbf{Encoder: \texttt{ViT-B/16}}& & & & \\
\midrule
Original$_{ViT-B/16}$ & \underline{65.82}$_{2.69}$ & \underline{28.28}$_{1.49}$ & 82.72$_{0.83}$ & \underline{42.98}$_{1.88}$ & \underline{68.63}$_{0.71}$ & 55.03$_{1.56}$ & $\underline{57.24}_{17.79}$ \\
TSplit$_{max}$ & 62.87$_{3.14}$ & 26.7$_{1.91}$ & \underline{83.7}$_{1.64}$ &40.19$_{2.6}$ & 68.43$_{1.53}$ & 55.24$_{1.62}$ & $56.19_{18.61}$ \\
TSplit$_{median}$ & 60.46$_{3.34}$ & 27.04$_{1.02}$ & 82.17$_{2.32}$ & 40.07$_{2.68}$ & 67.77$_{2.01}$ & \underline{56.54}$_{1.36}$ & $55.68_{17.96}$\\
TSplit$_{mean}$ & 63.96$_{3.24}$ & 25.92$_{0.73}$ & 83.38$_{1.93}$ & 39.88$_{1.21}$ & 67.61$_{1.9}$ & 56.18$_{0.75}$ & $56.16_{18.76}$\\
TSplit$_{percentile}$ & 61.88$_{1.25}$ & 26.13$_{0.83}$ & 82.62$_{1.33}$ & 41.3$_{1.82}$ & 67.6$_{2.6}$ & 54.84$_{1.65}$ & $55.73_{18.2}$\\
Baseline$_{ViT-B/16}$& 62.98$_{1.52}$ & 26.44$_{0.25}$ & 82.43$_{1.87}$ & 37.81$_{2.01}$ & 66.07$_{2.68}$ & 54.6$_{0.76}$ & 55.06$_{18.48}$\\
\midrule 
\end{tabular}
\end{adjustbox}
\caption{\textbf{Challenging setup}: \emph{TSplit uses the original test sets for evaluation and much less data for both training and model selection (see Table \ref{tab_tsplitsizes}).} \emph{Templateness}-based down-sampling compared against fine-tuning on the original dataset splits and against randomly downsampling to match the TSplit split sizes. We group results by their encoder (\emph{Original} subscript), where encoders are organized by size in descending order. The best performer in each group is \underline{underlined}, while the best for each dataset is in \textbf{bold}. See Table \ref{datasum} for evaluation measures. The \emph{Overall} column is the mean and standard deviation of each row for insight into aggregated performance.}
\label{tab_resamplecf}
\end{table*}
\label{resample_section}
\subsection{Down-sampling Experimental Setup} Here, we describe different experimental setups we investigated with TSplit. Our goal is to study the effect of meme templates on modelling, specifically on the classification datasets we have examined thus far. In these experiments we fine-tuned three different CLIP encoders. We opted for fine-tuning as prompting with (multimodal) LLMs does not result in meme understanding \cite{hwang-shwartz-2023-memecap}, and this is supported by our own experiments (see Section \ref{prompting}).


In this first suite of experiments, we down-sampled the training and validation data by using TSplit to reorganize them but kept the original test split untouched and used it for evaluation. In other words, we constructed a training, validation, and dummy test split from the original training and validation split based on their size ratios relative to the original sizes. We then used the resampled training data for fine-tuning and the resampled validation data for model selection, and discarded the dummy test split, using the original test data for the final evaluation. Therefore, less data was used in this setting for model optimization compared to the original data splits (see Table \ref{tab_tsplitsizes}).

We compare TSplit against two baselines: (1) CLIP fine-tuned on the original dataset splits (Original$_{ViT-X}$), using both OCR-extracted text and visual content, and (2) the same model with the training and validation data randomly downsampled to match TSplit's split sizes (Baseline$_{ViT-X}$; see Table \ref{tab_tsplitsizes}). Both baselines include a feed-forward layer after the CLIP text and image encoders. We also fine-tune the same model on our resplit datasets. Training is performed for 20 epochs using AdamW \cite{loshchilov2019decoupled} with a learning rate of $1e^{-5}$. Test evaluation uses the best-performing checkpoint on the validation split. For datasets without a validation split, we sample 20\% of the training data. We experiment over five seeds, reporting the mean and standard deviation.\footnote{We used 40 GB and 80 GB NVIDIA A100 Tensor Core GPUs.}

\subparagraph{Results and Discussion} Table \ref{tab_resamplecf} summarizes our experimental results. While our baselines performed strongly, TSplit$_{max}$ with the largest encoder outperformed them, achieving an F1 score 66.16 for MultiOff in the initial grouping, SOTA performance. For Memotion 3 (tasks A and B), performance remained consistent across configurations. Prior research highlights random (down)sampling as a strong baseline \cite{ein-dor-etal-2020-active, lu-etal-2024-strings}, likely due to the "noise" it introduces, which can enhance robustness.

Fine-tuning on the original split also established the state-of-the-art performance for FigMemes, achieving an F1 score of 47.79. However, this was subsequently surpassed by TSplit$_{max}$ using the largest encoder, which achieved an F1 score of 48.85. Notably, this dataset exhibited the greatest variance in model performance, with random downsampling proving to be comparatively less effective. We attribute this disparity to the inherent difficulty of the task and the highly templatic nature of the dataset (see Section \ref{eda}). The model benefits from either more data or template-based sampling, with TSplit demonstrating both sample efficiency and strong overall performance.

We observed that a variant of TSplit outperformed both of our baselines on four of the six datasets analyzed, while TSplit$_{max}$ from the first grouping emerges as the overall strongest performer. By accounting for \emph{templateness}, the model acquires more generalized meme knowledge that transfers effectively across splits, achieving this in an efficient manner. Notably, the sizes of the TSplit dataset splits used for optimization are significantly smaller than the original dataset (see Table \ref{tab_tsplitsizes}). Despite the reduction in data, this sampling approach matches or exceeds the performance of models trained on the full dataset. For FigMemes, as well as MAMI (tasks A and B), optimization was performed using nearly 1,000 fewer memes, yet this reduction led to improved performance.

Controlling for \emph{templateness} proves to be sample-efficient not only for datasets with a high proportion of templatic memes, such as FigMemes, but also for less templatic datasets like MAMI (see Section \ref{eda}). TSplit$_{max}$, when paired with the largest encoder, performed well on MultiOff, FigMemes, and MAMI (B), establishing itself as the overall strongest performer. TSplit$_{max}$ implies a high threshold for the maximum distance between a template and its examples. This is intuitive, as templatic memes can vary greatly from their base form (Figure \ref{retrival}), and even non-templatic memes may implicitly reference a template (Figure \ref{non_templatic}).

TSplit leverages the size and strong base performance of the encoder. Among both the baselines and TSplit configurations, \texttt{ViT-L/14@336px}—the largest encoder—demonstrates the strongest overall performance. Notably, random downsampling with this encoder can outperform optimization on the full dataset, which is markedly not the case for the other CLIP models. TSplit utilizes this encoder not only for classification but also to reorganize splits and selectively discard data. The model's strong performance suggests a latent understanding of meme content, which TSplit naturally leverages.

\begin{table*} [t]
\centering
    \begin{adjustbox}{width=.85\textwidth}
    \begin{tabular}{lccccccc}
    \toprule
    \textbf{Split} & \textbf{MultiOff} & \textbf{Memotion 3 (A)} &\textbf{Memotion 3 (B)} & \textbf{FigMemes} & \textbf{MAMI (A)} & \textbf{MAMI (B)}  & \textbf{Overall}\\
    \midrule
& & & \textbf{Encoder: \texttt{ViT-L/14@336px}}& & & & \\
\midrule
Original$_{ViT-L/14@336px}$ & \underline{63.64}$_{2.31}$ & 26.38$_{1.57}$ & \underline{81.7}$_{2.33}$ & 47.79$_{1.43}$ & 71.82$_{4.05}$ & 56.6$_{1.56}$ & 57.99$_{17.76}$ \\
TSplit$_{max}$ & 59.32$_{0.67}$ & 35.0$_{1.46}$ & 80.61$_{0.66}$ & \textbf{50.83}$_{1.37}$& \textbf{86.26}$_{0.67}$ & \textbf{61.18}$_{3.28}$
 & \textbf{62.2}$_{17.3}$\\
TSplit$_{median}$ & 57.03$_{2.88}$ & \textbf{35.3}$_{1.13}$ & 80.43$_{1.06}$ & 50.4$_{2.11}$ & 82.89$_{1.17}$ & 58.52$_{2.62}$ & 60.76$_{16.59}$\\
TSplit$_{mean}$ & 60.8$_{2.85}$ & 35.27$_{1.34}$ & 80.99$_{0.29}$ & 50.22$_{2.66}$ & 85.63$_{1.79}$ & 59.67$_{1.46}$ & 62.1$_{17.22}$ \\
TSplit$_{percentile}$ & 59.61$_{2.68}$ & 35.25$_{1.13}$ & 79.85$_{0.91}$ & 48.55$_{1.51}$ & 84.41$_{1.35}$ & 61.09$_{2.48}$ & 61.46$_{16.94}$\\
\midrule 
& & & \textbf{Encoder: \texttt{ViT-B/32}}& & & & \\
\midrule
Original$_{ViT-B/32}$ &  \underline{60.34}$_{2.21}$ & 28.46$_{1.67}$ & \textbf{82.65}$_{1.04}$ & 38.26$_{1.59}$ & 68.17$_{1.05}$ & 54.0$_{0.82}$ & 55.31$_{18.03}$ \\
TSplit$_{max}$ & 58.53$_{4.08}$ & \underline{32.13}$_{2.07}$ & 81.29$_{0.89}$ & \underline{40.98}$_{2.67}$ & 82.47$_{1.83}$ & 58.36$_{2.53}$ & \underline{58.96}$_{18.69}$ \\
TSplit$_{median}$ & 58.22$_{2.18}$ & 31.86$_{1.36}$ & 81.05$_{0.72}$ & 36.71$_{1.66}$ & \underline{83.47}$_{2.14}$ & 59.36$_{3.37}$ & 58.44$_{19.65}$ \\
TSplit$_{mean}$ & 57.42$_{4.11}$ & 31.35$_{1.92}$ & 80.97$_{0.55}$& 37.17$_{2.31}$ & 81.69$_{2.96}$ & \underline{59.61}$_{3.26}$ & 58.04$_{19.3}$\\
TSplit$_{percentile}$ & 54.57$_{6.21}$ & 30.91$_{1.93}$ & 79.84$_{0.88}$ & 39.24$_{3.95}$ & 83.09$_{1.59}$ & 56.45$_{2.9}$ & 57.35$_{19.17}$\\
\midrule 
& & & \textbf{Encoder: \texttt{ViT-B/16}}& & & & \\
\midrule
Original$_{ViT-B/16}$ & \textbf{65.82}$_{2.69}$ & 28.28$_{1.49}$ & \underline{82.72}$_{0.83}$ & 42.98$_{1.88}$ & 68.63$_{0.71}$ & 55.03$_{1.56}$ & 57.24$_{17.79}$ \\
TSplit$_{max}$ & 57.5$_{5.44}$ & 32.34$_{2.51}$ & 80.74$_{0.51}$ & 43.83$_{1.59}$ & \underline{83.93}$_{1.0}$ & 56.68$_{2.89}$ & 59.17$_{18.45}$ \\
TSplit$_{median}$ & 56.19$_{2.73}$ & \underline{33.11}$_{2.48}$ & 80.79$_{0.41}$ & \underline{43.85}$_{2.97}$ & 82.5$_{1.79}$ & 56.77$_{0.92}$ & 58.87$_{17.98}$\\
TSplit$_{mean}$ & 50.4$_{5.4}$ & 32.04$_{1.89}$ & 80.62$_{0.68}$ & 43.69$_{2.77}$ & 83.21$_{1.2}$ & 59.55$_{1.34}$ & 58.25$_{18.64}$\\
TSplit$_{percentile}$ & 58.52$_{5.41}$ & 31.42$_{2.1}$ & 80.54$_{0.82}$ & 43.04$_{2.91}$ & 82.85$_{1.34}$ & \underline{60.22}$_{3.06}$ & \underline{59.43}$_{18.48}$\\
\midrule 
\end{tabular}
\end{adjustbox}
\caption{\textbf{Easy setup:} \emph{TSplit resamples all data based on templatness, even resampling the test data.} We compare TSplit against fine-tuning and evaluating on the original dataset splits (see Table \ref{tab_wholetsplitsizes}).}
\label{tcf_table}
\end{table*}
\subsubsection{\textbf{Analysis:} TSplitting the Entire Dataset}
To better understand sampling with TSplit, we examine the strength of \emph{templateness} by applying TSplit and fine-tuning across the entire dataset. In the previous section, we observed that we are able to down-sample the training and validation splits with TSplit, and still fine-tune competitive models evaluated on the same test data. Here, however, we resample everything, comparing this model against Original$_{ViT-X}$. Table \ref{tab_wholetsplitsizes} shows an example of the TSplit split sizes across the entire dataset. Our fine-tuning procedure is the same.
\begin{table}[t]
\centering
    \begin{adjustbox}{width=0.45\textwidth}
    \begin{tabular}{lccccccc}
    \toprule
    \textbf{Dataset} & \textbf{Training Size} & \textbf{Validation Size} &  \textbf{Test Size}\\
     \midrule
     MultiOff & 473 & 119 & 151   \\
     Memotion 3 & 5700 & 1426 & 1374   \\
     FigMemes & 3246 & 362 & 1533  \\
     MAMI & 7996 & 1999 & 1005 \\ 
    \midrule
    \end{tabular}
    \end{adjustbox}
    \caption{Example TSplit dataset split sizes derived from the entire dataset. The encoder was \texttt{ViT-L/14@336px}.} 
    \label{tab_wholetsplitsizes}
\end{table}

\subparagraph{Results and Discussion}
Table \ref{tcf_table} shows our results. The TSplit versions of MultiOff appear more challenging than the original. Looking into the original test split revealed that it contained fewer templatic memes, whereas the TSplit test split included more. This suggests that templatic structures are harder for the model to learn in smaller datasets and that the original split was less challenging. 

For Memotion 3 (A), a difficult dataset with many templatic memes, previous work required the use of a ``Hinglish'' BERT-based model to reach an F1 of 33.28 in \citet{mishra2023memotion}. If we attempt to decouple the distribution from templates, we get multilingualism without even trying, attaining an F1 of 35.3 with TSplit$_{mean}$ from the first grouping.

 Controlling for template awareness makes difficult tasks easier by forcing the model to learn general, meme-specific properties. \citet{aggarwal-etal-2024-text} recently achieved an F1 of 85.0 on MAMI (A) using OCR text, focusing their investigation on hate speech in memes. Previously reported generic results in \citet{zhang-wang-2022-srcb} relied on fine-tuning and ensembling multimodal models, including CLIP (see Table \ref{datasum}), suggesting that both tasks are difficult, and our own results show that CLIP alone is a poor performer. While TSplit is based on the concept of meme templates, our UIs account for non-templatic memes or images, common in meme datasets. By removing meme/image conceptual overlap between the train and the test split, our approach is general and the model cannot rely on leaked information or spurious artifacts, resulting in two tasks that no longer require ensembling.

A tolerant view of template instances results in consistent, strong performance, especially with the largest encoder, where once more TSplit$_{max}$ is the overall strongest system, consistent with our earlier findings. The efficacy of TSplit$_{max}$ is corroborated not only by the results presented here but also by two additional experimental setups. In the first set of experiments, we investigated the regularization properties of TSplit. For this analysis, the models were fine-tuned over 20 epochs prior to the final evaluation and were therefore very likely overfit. TSplit$_{max}$, when paired with the largest encoder, once more emerged as the most robust performer overall. Furthermore, we conducted experiments involving the application of TSplit to the entire dataset, followed by downsampling the optimization data based either on \emph{templateness} or through random downsampling of the reorganized splits to match the size of the original downsampling experiments (see Table \ref{tab_tsplitsizes}). Notably, downsampling based on \emph{templateness} was capable of discarding up to 2,500 samples while achieving comparable, if not superior, performance relative to random downsampling or standard fine-tuning. These findings provide further evidence of the sample efficiency of meme templates (see Section \ref{moretcf}).

\begin{figure*}[t]
    \centering
    \includegraphics[width=.85\textwidth]{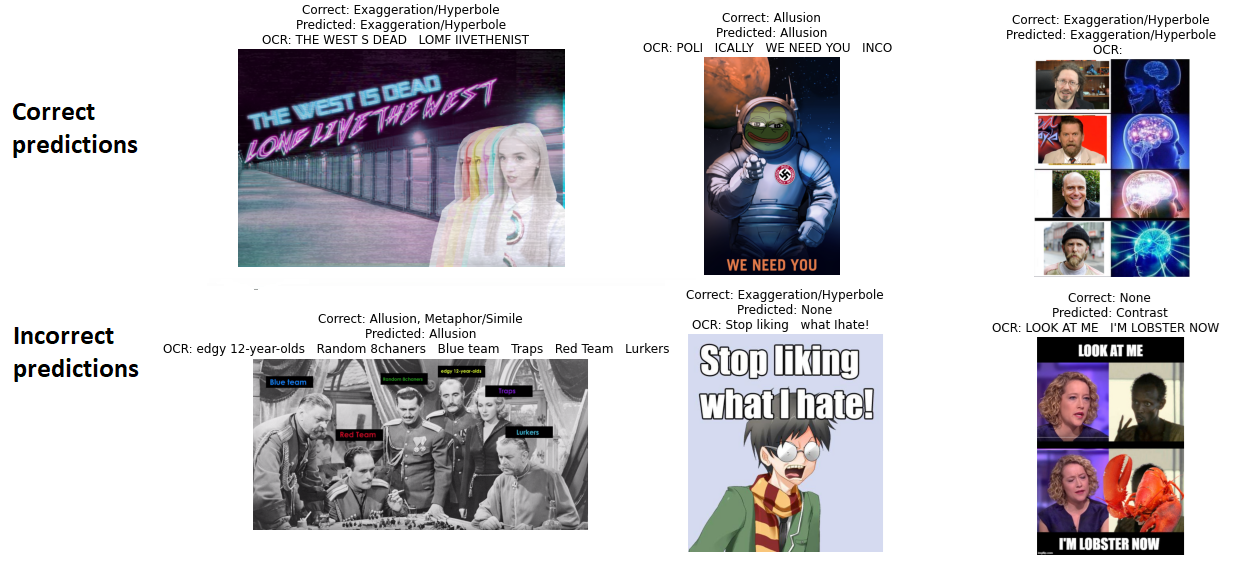}
    \caption{Representative examples of correct and incorrect predictions from TSplit on the FigMemes dataset.}
    \label{fig:error_analysis}
\end{figure*}

\subsection{TSplit Error Analysis}
We performed error analysis by examining correct and incorrect predictions made by TSplit on the FigMemes dataset. Note that we examined samples that were misclassified by Original$_{\texttt{ViT-L/14@336px}}$, i.e., difficult memes.

TSplit appears to be robust to errors in the extracted OCR text, as it predicted the correct label for the first and second entries in Figure \ref{fig:error_analysis}, despite the OCR text being incorrect and the memes being arguably non-templatic.
The model also excelled in cases involving templatic memes that follow established formats or memes that employ a templatic character, such as Pepe the Frog. For example, "Galaxy Brain"\footnote{\url{https://knowyourmeme.com/memes/galaxy-brain}} memes were accurately classified, demonstrating the model's strength in recognizing hyperbolic and exaggerated content within familiar templates. Similarly, the "Pepe Astronaut" meme was correctly classified as an allusion. TSplit's familiarity with memes like  Pepe made it robust to erroneous OCR text.

TSplit struggles with multilabel, non-templatic memes, such as the first entry in the second row, the model only predicted Allusion, missing the additional Metaphor/Simile label. Non-meme images also pose an issue, such as the second entry, which appears to be a cartoon referencing Harry Potter. A similar misclassification arose with the "Look at Me, I’m the Captain Now"\footnote{\url{https://knowyourmeme.com/memes/look-at-me-im-the-captain-now}} meme, classified as "Contrast" rather than "None". This template relies on juxtaposition for humor, but the model appeared to over-rely on structural cues, misinterpreting its intent. This may be due to the meme referencing an obscure Jordan Peterson interview\footnote{\url{https://www.youtube.com/watch?v=aMcjxSThD54}} and the template being heavily manipulated with overlaid images of Cathy Newman and a lobster.

The most challenging cases involved non-templatic memes or images that are arguably not memes. The World War II "Teams" meme illustrates this difficulty. While the model correctly identified "Allusion" based on the historical imagery, it failed to capture the metaphorical elements introduced by labels like "Blue Team," "Red Team", and "Lurkers." These labels transform the meme into a commentary on social dynamics of internet subcultures, adding a layer of metaphor that the model missed. Incorporating culturally and historically informed optimization data is an interesting topic we leave for future research.

\section{Conclusion and Future Work}
We analyzed the power of meme templates. To anchor our analysis to memes, we created the KYMKB, containing more than 54,000 images and 5,200 base templates with detailed information about each one. We first showed that a comparison of templates to memes creates a strong signal that we can leverage in downstream tasks. We therefore proposed TSplit and found it is sample-efficient and can result in robust models. It was not our goal to create state-of-the-art meme classifiers, but we believe our methods are convincing demonstrations of the strength of meme templates -- memes may be difficult to analyze, but they are not random.

The potential applications of meme templates are vast. Our findings demonstrate their significant impact on topics such as retrieval (see Section \ref{eda}), sample efficiency, model performance, and even multilingual capabilities. However, our work represents only an initial exploration of this area. Future research could further investigate these applications and other critical directions, contributing to a deeper understanding of the role and utility of meme templates in computational research.

Our resources provide unified, inexpensive tools for future research grounded in meme knowledge, and specifically, meme templates. In future work, we will apply the KYMKB to more datasets and languages in a cross-language setup and explore automatically augmenting the KYMKB with new memes and with new templates.

\section*{Acknowledgments}
We would like to thank Tobin Bates, Max Glockner, Derek Hommel, and Timour Igamberdiev for sharing inspiring discussions with us and Irina Bigoulaeva and Seolhwa Lee for their invaluable feedback on an early draft of our manuscript. LB and IG acknowledge the German Federal Ministry of Education and Research (BMBF) under the promotional reference 13N15897 (MISRIK) and the Hessian Ministry of Science and the Arts (HMWK) within the projects "The Third Wave of Artificial Intelligence - 3AI" (200009-62200024), hessian.AI and within their joint support of the National Research Center for Applied Cybersecurity ATHENE. PEC was supported by the Pioneer Centre for AI, DNRF grant number P1.

\section*{Limitations}
KYM is in our view the best resource for meme-related knowledge, but this does not mean that it is the only resource, nor does it mean that all meme posters necessarily agree on the interpretation of a template or a meme. Like all forms of communication, there is ambiguity in what a given instance means. Not all memes are templatic, but it is our belief that the most popular memes are, at least based on how meme datasets are created. TSplit quantifies the notion of templateness for the sake of computation, however, we believe that truly determining the templateness of a given meme is not trivial and it is certainly not the case that KYM contains all known templates. We have devised a measure by which to determine templateness, but it is only applicable within the limited scope of ML, where memes are viewed as images (see Section \ref{whatmeme}). We have performed an examination of multimodal LLM prompting performance and found such a paradigm to be insufficient for meme understanding (see Section \ref{prompting}) and our findings are consistent with \citet{hwang-shwartz-2023-memecap}. More recent vision language models, such as Otter \cite{li2023otter} or IDEFICS,\footnote{\url{https://huggingface.co/blog/idefics}} might be stronger, we are skeptical due to their incremental nature and the poor performance of LLaVA \cite{liu2023improved}, even supplemented by the KYMKB. However, we did not test this ourselves. We have not considered pay-to-use corporate artifacts for reasons of reproducibility and accessibility.

While our analysis demonstrates that grounding in templates can have multilingual effects, we have not addressed the cultural or linguistic limitations or potential biases inherent in our approach or our knowledge base. Future work should include an in-depth investigation of KYMKB's diversity in terms of culture, language, and humor.

There is an argument to made for template instances being equally distributed between dataset splits, as a model has no hope of interpreting a novel template at test time. This is not our own view because it is our belief that we should be testing model robustness on memes that are as novel as possible. However, we enable this functionality in TSplit to encourage future research on model-template understanding.

The distinction between different meme types is not always clear and is arguably subjective. In future work, we will use the KYMKB to develop a taxonomy of memes in order to aid the development of meme-aware systems.

\section*{Ethics Statement}
It is possible that the resources and insights we have developed and discussed may be misused to spread harmful memes more effectively. Abuse is an unfortunate drawback of all tools and technology. We hope that our work is used instead to create systems that have stronger meme understanding such that we can automatically and accurately flag dangerous memes to halt their spread on social media. To this end, we created and demonstrated how the KYMKB can be used practically and analytically. To investigate model-meme understanding, we conducted thorough classification experiments with adversarial meme datasets via TSplit and found that we can use it to discourage models from taking undesirable shortcuts to meme understanding, which results in more robust models. Throughout this work, we have emphasized the dangers that memes can present, pointed out how our field is lacking in its approach to memes, and taken what we believe are the first steps on a long road to intelligent systems that understand memes.

We do not publish data, but following similar work that provided access to real-world content \cite{fan-etal-2019-eli5, hanselowski-etal-2019-richly, zlatkova-etal-2019-fact}, we make our scraping and organization code available to recreate the KYMKB and use Wayback Machine\footnote{\url{https://wayback-api.archive.org/}} (WM) as a best effort to ensure reproducibility, as the WM URLs will not foreseeably change. If KYM does not wish for their content to be accessed via WM, they have the option to opt out,\footnote{\url{https://help.archive.org}} such that their content is removed.

\bibliography{new_custom}
\bibliographystyle{acl_natbib}

\appendix

\section{Appendix}
\label{sec:appendix}
\begin{table*} [t]
\centering
    \begin{adjustbox}{width=.85\textwidth}
    \begin{tabular}{lccccccc}
    \toprule
    \textbf{Split} & \textbf{MultiOff} & \textbf{Memotion 3 (A)} &\textbf{Memotion 3 (B)} & \textbf{FigMemes} & \textbf{MAMI (A)} & \textbf{MAMI (B)} & \textbf{Overall} \\
    \midrule
     & & & \textbf{Encoder: \texttt{ViT-L/14@336px}} & & & & \\
    \midrule
Original$_{ViT-L/14@336px}$ & $59.77_{3.14}$ & $26.77_{1.04}$ & $\underline{79.8_{0.85}}$ & $44.93_{3.03}$ & $66.72_{1.22}$ & $51.93_{1.18}$ & $54.99_{16.75}$ \\

TSplit$_{max}$ & 62.16$_{2.14}$ & 32.96$_{1.26}$ & 77.95$_{1.04}$ & \textbf{47.91$_{1.57}$} & \textbf{83.63$_{2.36}$} & \textbf{59.65$_{4.66}$} & \textbf{60.7}$_{17.12}$\\
TSplit$_{median}$ & 58.27$_{2.57}$ & 34.73$_{1.15}$ & 78.49$_{0.97}$ & 42.51$_{9.36}$ & 82.15$_{2.33}$ & 56.33$_{1.4}$ & 58.75$_{17.24}$\\
TSplit$_{mean}$ & 61.57$_{3.98}$ & \textbf{35.04$_{1.46}$} & 78.1$_{1.74}$ & 45.36$_{3.5}$ & 81.22$_{2.62}$ & 56.7$_{3.35}$ & 59.66$_{16.47}$\\
TSplit$_{percentile}$ & \underline{63.58$_{3.0}$} & 33.53$_{2.59}$ & 77.38$_{1.94}$ & 40.26$_{10.87}$ & 80.52$_{0.92}$ & 53.32$_{2.93}$ & 58.1$_{17.56}$\\
\midrule 
 & & & \textbf{Encoder: \texttt{ViT-B/32}} & & & & \\
    \midrule
Original$_{ViT-B/32}$ & \underline{60.43$_{1.63}$} & 29.08$_{0.9}$ & \textbf{80.26$_{1.3}$} & 35.92$_{2.34}$ & 66.06$_{1.99}$ & 51.97$_{1.23}$ & 53.95$_{17.46}$\\
TSplit$_{max}$ & 56.02$_{2.38}$ & 33.2$_{1.27}$ & 78.04$_{1.27}$ & 37.96$_{1.52}$ & 79.74$_{1.42}$ & 55.53$_{3.04}$ & 56.75$_{17.76}$ \\
TSplit$_{median}$ & 53.19$_{4.32}$ & 33.92$_{1.64}$ & 77.8$_{0.87}$ & \underline{39.8$_{4.32}$} & 79.87$_{1.81}$ & \underline{56.08$_{2.38}$} & $\underline{56.92}_{18.97}$\\
TSplit$_{mean}$ & 57.73$_{3.44}$ & 32.54$_{1.39}$ & 76.84$_{0.55}$ & 36.68$_{1.71}$ & \underline{80.13$_{1.68}$} & 55.16$_{2.61}$ & $56.51_{17.99}$\\
TSplit$_{percentile}$ & 55.68$_{4.92}$ & \underline{34.49$_{1.7}$} & 76.41$_{2.2}$ & 38.25$_{2.72}$ & 77.17$_{3.65}$ & 55.31$_{1.94}$ & $56.22_{16.55}$\\
\midrule 
 & & & \textbf{Encoder: \texttt{ViT-B/16}} & & & & \\
    \midrule
Original$_{ViT-B/16}$ & \textbf{64.65$_{2.12}$} & 27.28$_{0.65}$ & \underline{79.49$_{3.0}$} & 40.58$_{2.0}$ & 67.61$_{1.96}$ & 51.5$_{2.65}$ & $55.18_{17.51}$\\
TSplit$_{max}$ & 58.23$_{4.87}$ & \underline{34.95$_{1.42}$} & 77.47$_{0.64}$ & 39.4$_{4.93}$ & 80.1$_{0.83}$ & \underline{57.13$_{4.11}$} & $57.88_{17.06}$\\
TSplit$_{median}$ & 57.31$_{2.91}$ & 33.33$_{1.54}$ & 78.34$_{1.59}$ & 40.96$_{0.7}$ & \underline{80.74$_{1.29}$} & 54.36$_{2.69}$ & $57.51_{17.52}$\\
TSplit$_{mean}$ & 59.74$_{3.2}$ & 34.16$_{2.64}$ & 77.28$_{1.83}$ & 40.16$_{3.19}$ & 80.31$_{1.21}$ & 55.99$_{3.38}$ & $57.94_{17.14}$\\
TSplit$_{percentile}$ & 59.4$_{6.11}$ & 34.5$_{1.89}$ & 77.95$_{1.34}$ & \underline{42.71$_{2.02}$} & 79.28$_{0.71}$ & 54.33$_{3.45}$ & $\underline{58.03}_{16.95}$\\
\midrule 
\end{tabular}
\end{adjustbox}
\caption{TSplit compared against the original dataset. We group results by their encoder (\emph{Original} subscript), where the encoders are organized by size in descending order. The best performer in each group is \underline{underlined}. The best performer for each dataset is in \textbf{bold}. The \emph{Overall} column is the mean and standard deviation of each row for insight into performance on aggregate.}
\label{old_tcf_table}
\end{table*}
\subsection{What's in a Meme?}
\label{whatmeme}
Memes are not just images that sometimes have text. The KYMKB captures this fact and how far we as a community are from meme understanding. Consider \emph{Leeroy Jenkins},\footnote{\url{https://knowyourmeme.com/memes/leeroy-jenkins}} a template that references a popular YouTube video\footnote{\url{https://www.youtube.com/watch?v=mLyOj_QD4a4&t=1s}} where a player in World of Warcraft\footnote{\url{https://worldofwarcraft.blizzard.com}} makes a brash decision while yelling his name, Leeroy Jenkins. This results in a party of players losing a fight to a monster.

An instance of this template is not merely some image, but rather hollering \emph{Leeroy Jenkins} or using the audio from the original template when performing a reckless act that will likely have negative consequences. A concrete example of this can be seen in a recent YouTube video.\footnote{\url{https://www.youtube.com/watch?v=UdWv202brqo} (at 1:25).} We are unaware of any approach which considers memes in audio form. Despite this template originating in 2005, it is still referenced almost 20 years later, demonstrating the longevity of popular templates. The video in question is a compilation of memes, but is not composed of still images sometimes with text, but rather audio and video. At the time of writing, this video has more than 6.6 million views, which we feel is compelling evidence that this is a more realistic representation of memes than what can be found in the literature. This video is not an edge case either, but rather a case that has not been considered in previous work, exemplified by the relevant YouTube channel having 18 other such videos, each with more than one million views. 
Such examples may seem anomalous, but we argue otherwise and we believe that such an interpretation is a consequence of the narrow scope of the literature. In Section \ref{edging}, we provide a detailed discussion about additional \emph{edge case} examples contained within the KYMKB.

In order to make our work digestible, we have conformed to the notion of memes that the AI community has converged to. TSplit, for example, relies on the concept that memes are images in order to perform classification, but our method is meant to demonstrate the usefulness of templates and a shortcoming of the literature. Templatic memes are only the tip of the iceberg when it comes to understanding this form of communication and the KYMKB provides a wealth of knowledge we can utilize to create systems capable of interpreting memes.

\subsection{Template-Aware Splitter details}
\label{tsplit_deets}
\label{moretcf}
In this Appendix section, we provide additional results from fine-tuning with TSplit. Our experimental conditions are the same as Section \ref{tsplit} except here we perform inference with the model fine-tuned for $20$ epochs to highlight TSplit's regularization effects. These results can be found in Table \ref{old_tcf_table}.

In the case of MultiOff, we see that smaller encoders struggle with our resplit datasets, while the larger encoder overfits to the original split. In such cases, TSplit appears to have a regularization effect. 
For Memotion 3 (A), a difficult task with many 
templatic memes, TSplit outperforms our baseline and consistently attains scores of approximately 33–35, and we again get multilingualism without even trying. Task B tells a different story, where the TSplit datasets appear harder than the original.

FigMemes remains a challenging task, and the largest models are required to be performant, but are still prone to overfitting. We again note the regularization properties of TSplit, beating our baseline with an F1 of 47.91.

\begin{table*}[t]
\centering
    \begin{adjustbox}{width=\textwidth}
    \begin{tabular}{lccccccc}
    \toprule
    \textbf{Dataset} & \textbf{ViT-B/16 Training Size} & \textbf{ViT-B/32 Training Size} & \textbf{ViT-B/16 Validation Size} & \textbf{ViT-B/32 Validation Size} & \textbf{ViT-B/16 Dummy Test Size} & \textbf{ViT-B/32 Dummy Test Size}\\
     \midrule
     MultiOff & 367 & 354 & 93 & 90  & 134 & 150   \\
     Memotion 3 & 4930 & 4723  & 1233 & 1181 &  837 & 1096   \\
     FigMemes & 2293 & 2327 & 256 & 260 & 1050 & 1012  \\
     MAMI & 7213 & 7295 & 1804 &  1824  & 983 & 881 \\ 
     \midrule
    \end{tabular}
    \end{adjustbox}
    \caption{Example TSplit dataset split sizes for \texttt{ViT-B/16} and \texttt{ViT-B/32}. The dummy test data was discarded.}
    \label{tab_tsplitsizes_encoder}
\end{table*}

\subsubsection{TSplitting the Enitre Dataset and Downsampling}
In this section, we present additional experiments to further investigate the efficacy of TSplit sampling. Specifically, we resample the entire dataset using TSplit, followed by downsampling and selective discarding of optimization data in two distinct ways: 1) by randomly downsampling to match the sizes outlined in Table \ref{tsplit_splits}, and 2) by sampling templates that may appear in the training or validation data based on dividing the random downsampled size by the original training size, i.e., \( \text{train\_ratio} = \frac{\text{downsample\_size}}{\text{original\_training\_size}} \).
 As detailed in Section \ref{tsplit}, we accomplish this by generating an array of uniquely detected templates and unique identifiers (UIs) that are eligible to appear in the training data. This array is then shuffled, and a cutoff index is determined. All detected templates or UIs preceding the cutoff index is discarded, while the remaining data is used for training. Once we have the ratio, we then multiply it by the number of detected training templates, as follows:

\[
\text{cutoff} = \left\lfloor \text{train\_templates} \times \text{train\_ratio} \right\rfloor
\]

Similarly, to create the validation data, we sample from the training data based on the ratio of the randomly downsampled split size to the size of the original split. This is an aggressive means of downsampling the optimization data based on templateness, where we might discard more then $2,500$ samples (see Table \ref{tab:tsplit_style_downsampling}).

\begin{table}[ht]
\centering
\small
\begin{tabular}{lcccc}
\toprule
\textbf{Dataset} & \textbf{Training} & \textbf{Validation} & \textbf{Test} & \textbf{Discard} \\
\midrule
MultiOff    & 461   & 69    & 137   & 76   \\
Memotion3   & 3650  & 913   & 1609  & 2328 \\
FigMemes    & 2683  & 143   & 1376  & 939  \\
MAMI        & 6145  & 1537  & 740   & 2578 \\
\bottomrule
\end{tabular}
\caption{Dataset sizes for MultiOff, Memotion3, FigMemes, and MAMI with full TSplit downsampling.}
\label{tab:tsplit_style_downsampling}
\end{table}

Our experimental setup remains the same, but we only experimented with the \texttt{ViT-L/14@336px} CLIP model. We compare these two downsampled TSplit versions against fine-tuning on the original splits.

\begin{table*} [t]
\centering
    \begin{adjustbox}{width=.85\textwidth}
    \begin{tabular}{lccccccc}
    \toprule
    \textbf{Split} & \textbf{MultiOff} & \textbf{Memotion 3 (A)} &\textbf{Memotion 3 (B)} & \textbf{FigMemes} & \textbf{MAMI (A)} & \textbf{MAMI (B)} & \textbf{Overall} \\
    \midrule
Original$_{ViT-L/14@336px}$ & \textbf{63.64}$_{2.31}$ & 26.38$_{1.57}$ & \textbf{81.7}$_{2.33}$ & 47.79$_{1.43}$ & 71.82$_{4.05}$ & 56.6$_{1.56}$ & $57.99_{17.76}$ \\
\textit{Random Downsampling}\\ 
TSplit$_{max}$ & 60.21$_{1.9}$ & 35.12$_{1.0}$ & \underline{80.74}$_{0.62}$ & 48.34$_{3.17}$ & 84.32$_{3.36}$ & 61.45$_{2.88}$ & \underline{61.7}$_{17.13}$\\
TSplit$_{median}$ & 58.18$_{7.07}$ & 33.13$_{2.03}$ & 80.34$_{0.45}$ & \textbf{50.15}$_{2.19}$ & 82.09$_{1.73}$ & 59.3$_{2.9}$ & $60.53_{16.95}$\\
TSplit$_{mean}$ & 59.91$_{5.28}$ & \underline{35.51}$_{0.81}$ & 80.44$_{0.47}$ & 47.2$_{0.98}$ & 83.15$_{1.74}$ & \underline{61.65}$_{2.63}$ & $61.31_{16.88}$\\
TSplit$_{percentile}$ & \underline{62.59}$_{4.64}$ & 35.48$_{1.04}$ & 80.66$_{0.42}$ & 44.56$_{9.82}$ & \underline{84.94}$_{1.55}$ & 60.2$_{2.56}$ & $61.4_{17.72}$\\
\textit{TSplit Downsampling}\\ 
TSplit$_{max}$ & 58.83$_{2.05}$ & 33.77$_{1.35}$ & \underline{80.8}$_{0.88}$ & 46.19$_{2.02}$ & \textbf{85.71}$_{2.53}$ & 61.62$_{2.54}$ & $61.15_{18.11}$\\
TSplit$_{median}$ & 60.14$_{3.57}$ & \textbf{35.68}$_{2.06}$ & 80.36$_{0.62}$ & \underline{48.45}$_{2.63}$ & 83.85$_{1.75}$ & \textbf{63.05}$_{3.18}$ & $\textbf{61.92}_{16.81}$\\
TSplit$_{mean}$ & \underline{60.79}$_{3.81}$ & 33.48$_{1.6}$ & 79.84$_{0.43}$ & 46.35$_{2.09}$ & 84.15$_{0.45}$ & 59.39$_{1.87}$ & $60.67_{17.63}$\\
TSplit$_{percentile}$ & 60.18$_{4.35}$ & 35.24$_{1.44}$ & 80.59$_{0.92}$ & 47.55$_{1.26}$& 83.81$_{2.4}$ & 60.66$_{2.06}$ & $58.1_{17.56}$\\
\midrule 
\end{tabular}
\end{adjustbox}
\caption{Fine-tuning on the original splits is compared against T-Splitting the entire dataset followed by random downsampling, and T-Splitting the entire dataset with downsampling based on templateness. The best performer within each group is \underline{underlined}, and the best performer for each dataset is \textbf{bolded}. The \emph{Overall} column represents the mean and standard deviation for each row, providing insight into the aggregate performance.}
\label{tsplit_sampling}
\end{table*}

The results presented in Table \ref{tsplit_sampling} reveal several notable trends across datasets and sampling methods. Fine-tuning on the original splits achieves the strongest performance for MultiOff (\( \mathbf{63.6} \)) and Memotion 3 (B) (\( \mathbf{81.7} \)). 

Template-based downsampling demonstrates notable improvements, even for datasets that do not contain a high proportion of templates. For example, TSplit$_{\text{max}}$ from the last grouping achieves the best performance for MAMI (A) (\( \mathbf{85.71} \)). Similarly, TSplit$_{\text{median}}$ stands out as the top performer for MAMI (B) (\( \mathbf{63.05} \)) and achieves the highest overall mean score (\( 61.92\)), highlighting its robustness across datasets.

Interestingly, for FigMemes, the improvements with template-based downsampling are marginal, but we see a boost if we first TSplit the entire dataset and combine this with random downsampling. In contrast, for Memotion 3 (A), TSplit$_{\text{median}}$ (\( 35.51\)) performs best, underscoring that the optimal sampling method may vary depending on the dataset's characteristics.

Overall, template-based TSplit methods tend to outperform random downsampling or even using more data for optimization, that is, standard fine-tuning. These findings indicate that while fine-tuning on the original splits remains competitive for certain datasets, template-based sampling offers a more sample-efficient and effective alternative.

\subsubsection{Template-Split Analysis}

\begin{table}[ht]
     \centering
    \begin{adjustbox}{width=.45\textwidth}
    \begin{tabular}{lcccc}
    \toprule
    \textbf{Split and threshold} & \textbf{MultiOff} & \textbf{Memotion 3} & \textbf{FigMemes} & \textbf{MAMI}  \\
    \midrule
Train$_{max}$ & 228 / 4 & 1407 / 141 & 1335 / 59 & 1961 / 136 \\
Test$_{max}$ & 55 / 2 & 303 / 28 & 568 / 29 & 198 / 11  \\
Train$_{median}$ & 204 / 88 & 1170 / 2051 & 1155 / 547 &  1764 / 1191 \\
Test$_{median}$ & 50 / 23 & 249 / 441 & 525 / 204 & 172 / 123  \\
Train$_{mean}$ & 194 / 100 & 1106 / 2614 & 1122 / 655 &  1700 / 1510 \\
Test$_{mean}$ & 47 / 26 & 228 / 569  & 497 / 264 & 170 / 150 \\
Train$_{percentile}$ & 167 / 176 & 890 / 3562 & 964 / 1198  & 1506 / 2789  \\
Test$_{percentile}$ & 41 / 44 & 205 / 748 &  419 / 506 & 138 / 291\\
\midrule
\end{tabular}
\end{adjustbox}
\caption{Example of how TSplit reorganizes datasets, where we show the number of detected templates / number of unique identifiers in each split for each thresholding method. \texttt{ViT-L/14@336px} was used as the CLIP encoder.}
\label{tsplit_splits}
\end{table}
 
In this section, we examine how TSplit samples templates and unique identifiers. As Table \ref{tsplit_splits} shows, using the maximum distance from template to examples as the threshold value for each template results in TSplit detecting more templates, while using the $25^{th}$ percentile results in more unique identifiers.

To empirically verify how TSplit resplits datasets, we sampled 1000 memes from FigMemes from both the training and test split of our reorganized datasets. We then manually inspected the memes, looking for overlapping templates, template references, or meme-related artifacts. We did this under all four thresholding techniques, where the CLIP encoder was \texttt{ViT-L/14@336px}.

Under TSplit$_{max}$ and TSplit$_{median}$, we observed that certain templatic characters, such as Pepe the Frog, appeared in both splits. Pepe was present in another base template as well as in a non-templatic meme, but no single template appeared across both splits. In contrast, under TSplit$_{mean}$, we noted similar templatic characters and observed one overlapping template, Picardía / Thumbs Up Emoji Man, which is frequently used on 4chan to mock political ideologies.\footnote{\url{https://knowyourmeme.com/memes/picardia-thumbs-up-emoji-man}} This template’s flexibility in appearance suggests that a strict interpretation of template instances would naturally result in its presence in both splits. Under TSplit$_{percentile}$, while no explicit template overlap was noted, the same templatic characters—such as Pepe, Picardía, SpongeBob, and Spiderman—appeared across both splits.\footnote{\url{https://knowyourmeme.com/memes/spider-man-pointing-at-spider-man}} Although these examples are not instances of a formal template, they evoke a similar emotional resonance, akin to the templates they reference.

\subsection{Template-Label Counter}
\begin{figure*}[t]%
    \centering
    \includegraphics[width=.85\textwidth]{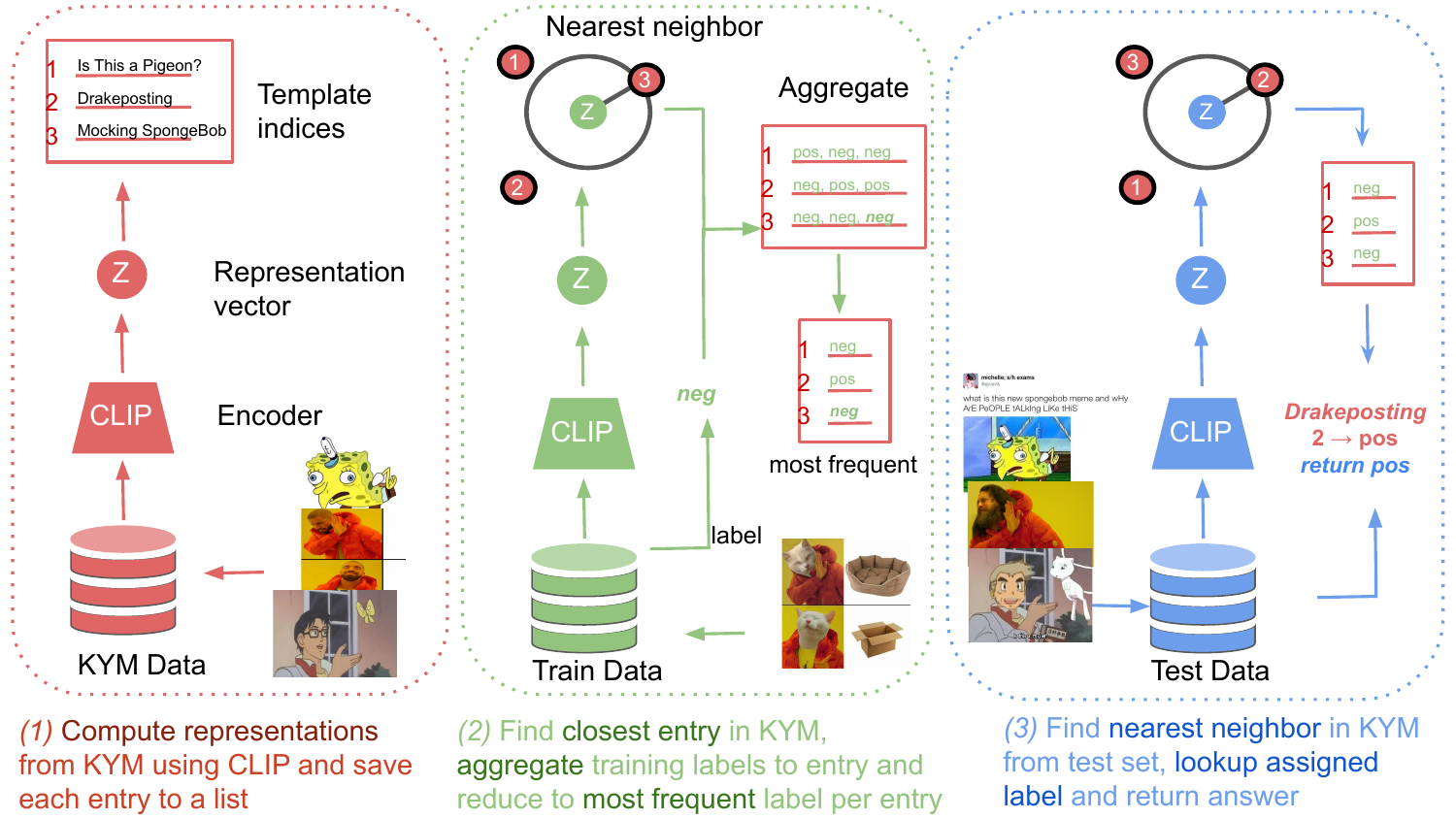}
    \caption{TLC encodes the KYMKB and computes a nearest neighbor index. We then encode the training data and query our lookup, recording each template's most frequent class. We query the index and assign the closest template's label to test samples. As an example, we use \emph{pos} and \emph{neg} as labels for a sentimental analysis task.}
    \label{tlc}
\end{figure*}
\label{tlc_original} 

We hypothesize that many meme datasets are often nothing more than examples of popular templates we have collected in the KYMKB. We should therefore be able to compare memes to templates, select the most similar template, and obtain a meme-specific context. To test this, we matched templates to memes in the training split of a dataset. We can then assign a meme's label to another meme if they share the same template, i.e.,~a novel meme in the test split of that dataset (see Figure \ref{tlc}).

\subparagraph{Meme knowledge} We again opted for nearest neighbor indexing as a similarity measure and formalize this as a ranking task, where we first create a template reference, $ref = f(X_{KYMKB})$. 

\subparagraph{Dataset knowledge}  To learn a dataset's labeling scheme, we encode the training data, $query_{train} = f(X_{train})$, and query our index, selecting the closest template and recording the label for each training instance. TLC then reduces each index (template) to the most frequent label: \[\operatorname*{arg\,max}_{ref} count(rank(ref, query_{train})) \] Here our $rank$ function sorts entries in the KYMKB in ascending order based on their Euclidean distance from a query vector.

\subparagraph{Testing meme and dataset knowledge} The final step is to encode test data, $query_{test} = f(X_{test})$, and then query our lookup. We then assign the most frequent label for a template to a test instance, $\hat{y} = rank(ref, query_{test})$. If we find a template not seen during training, we backoff to the most frequent label in the training data.

\subparagraph{Hyperparameter values} TLC has the option to ignore the meme itself and instead to match the \textit{about} section of templates to the OCR text of a novel meme. Alternatively, we can choose to consider base templates or also examples for encoding knowledge about the meme. Multiple neighbors can be searched over, selecting the most common template or label among them. We can also use multiple modalities, combining the \textit{about} section from the template/example and the OCR text, respectively, with the template and the novel meme embeddings. We experimented with concatenating the CLIP embeddings of both modalities, fusing via the Hadamard product, normalizing and averaging the two modalities as the final input vector \cite{meta-fusion}, and a type of late fusion, where the text and the image representations vote separately and we then aggregate. After the hyperparameter values are set, TLC is deterministic (see Appendix \ref{tlc deets}). Note that TLC is reliant on the KYMKB.

\subsection{Classification Experiments}
\label{experiments}

\begin{table*}[ht]
\centering
    \begin{adjustbox}{width=.85\textwidth}
    \begin{tabular}{lccccccc}
    \toprule
    \textbf{Method} & \textbf{MultiOff} & \textbf{Memotion 3 (A)} & \textbf{Memotion 3 (B)} & \textbf{FigMemes} & \textbf{MAMI (A)} & \textbf{MAMI (B)}\\
    \midrule
     Majority & 37.92 & 21.5 & 72.59 & 5.72 & 33.33 & 18.2 \\
     \midrule
     Best previous: text only & \textit{54.0} & NA & NA & 34.06  & 83.0 & NA  \\
     Best previous: vision only & 24.0 & NA & NA & \textit{47.69}  & \textit{85.0} & NA   \\
     Best previous: vision+text & 50.0 & \textit{33.28} & \textit{74.74} & 46.69 & 83.4 & \textit{73.1}   \\
     \midrule
     TLC$_{Text}$ & 51.83 & 35.4 & 77.6 & 21.14  & 61.86 & 35.93 \\ 
     TLC$_{Templates}$ & \textbf{61.89} & \textbf{37.77}  & 79.89 & \textbf{29.8}  & 69.24 & 39.99 \\
     TLC$_{Templates+Instances}$ & 58.58 & 37.04 & \textbf{80.49} & 28.97 & \textbf{70.0} &  \textbf{40.21} \\
     \bottomrule
    \end{tabular}
    \end{adjustbox}
    \caption{Classification results for the best-performing version of TLC (\textbf{in bold}) compared against the best performing method from the related work (\textit{in italics}). \emph{Instances} refers to template examples in the KYMKB. See Section \ref{more_clf_results} for TLC hyperparameter configurations and Appendix \ref{dataset_fino} for modelling information from previous work.}
    \label{classification}
\end{table*}
\noindent


\subparagraph{Baselines and experimental setup} We test various versions of TLC on six meme classification datasets: FigMemes with seven labels of different types of figurative language, MultiOff a binary dataset where the labels can be offensive or non-offensive, Memotion 3, where Task A has three labels for sentiment analysis and Task B has four for different types of emotion, and MAMI with Task A being  binary misogyny detection and Task B having four labels with different types of misogyny being expressed (see Tables~\ref{datasum} and ~\ref{modelsum}). We choose these datasets because they were made to address the issue of harmful meme detection and to frame our analysis of meme templates in the light of this important goal. Our baseline is a majority class classifier. Each dataset used Google Vision to provide OCR text overlaid on memes.\footnote{\url{https://cloud.google.com/use-cases/ocr}}
\label{discuss}

\subparagraph{Results and Discussion} Table \ref{classification} shows our results. We display the best-performing version of TLC, comparing embedding text versus templates versus templates and examples. We also show the best result from previous work, where a PLM was fine-tuned on OCR text, the meme itself, or a multimodal representation of the two. TLC beats our majority class classifier, but this baseline is competitive a fine-tuned PLM for Memotion 3 (B).


TLC's performance consistently improves as we consider more modalities. Encoding the \textit{about} section of a template and the OCR text from a novel meme is strong on its own, especially in the case of Memotion 3. As we add template and meme images, the performance improves by more than ten points for MultiOff. We find that concatenating the image and the text modalities tends to be the strongest TLC configuration, supporting our hypothesis that the base semantics of a meme is explained in the \textit{about} section, but is also captured by the template. We can naturally obtain a better representation of the exact meaning by using meme-specific information from OCR.


The base template is sufficient to encode meme knowledge and is more efficient than also embedding examples. For MultiOff, we see a boost of more than two points when we only consider templates. In other cases, TLC$_{Templates}$ is within one point if not higher than TLC$_{Templates+Instances}$. Encodng one-tenth of the available images results in a strong model, demonstrating that meme datasets can be instances of the KYMKB templates.

In the case of Memotion 3 and MultiOff, our approach is a stronger method than fine-tuning a PLM. We further note that meme templates cross cultural and linguistic boundaries, as indicated by our strong performance on both Memotion 3 datasets, a multilingual dataset of memes in Hindi and English. Templates give us multilinguality for free. 

TLC assumes memes belong to a template, but our prediction has no meaning for a picture (which is not a meme). Many meme datasets are not curated to remove non-memes, containing both memes and images. This can be verified in the datasets or by reading the paper. Figure 1 in FigMemes shows an example of a visual metaphor/simile, which is a picture, not a meme (see (f)). 

In Figure 1 of MAMI, all examples are not templatic memes and are understandable without knowledge of memes (see Figure \ref{non_templatic}).This is supported by our template-meme analysis, where we saw that a large percentage of memes in MAMI are unrelated to templates (see Section \ref{eda}).

\subsection{Template-Label Counter details}
\label{tlc deets}
In this Appendix section, we provide additional details about TLC that could not be provided in the main text due to space limitations. For TLC, each template is associated with an array of observed training labels derived from instances of that template. This array is then reduced to the most frequently occurring label, which serves as the template’s representative label. When the same template appears again, such as in the test data, this representative label is assigned to any novel memes associated with the template. There are actually multiple ways we can go about voting if we consider multiple neighbors. First, we could consider multiple templates and then take their most common label, only keeping and recording that label. We refer to this as \emph{template vote}. In cases where we only consider templates and not examples, this would mean often backing off to the most majority class in the dataset because we will find distinct templates. Alternatively, we could keep all labels for a given template and then reduce to its most frequent label, which we refer to as \emph{label vote}. We consider all cases. We find that the template style of voting is the strongest and it is about this configuration that we report results. The only exception to this is MAMI, where we found \emph{label vote} to be the best configuration. This finding is intuitive because MAMI is composed largely of memes which are not templatic and therefore it is the label signal, not the template signal, which is most beneficial for classification.
\begin{table*}[t]
\centering
    \begin{adjustbox}{width=.85\textwidth}
    \begin{tabular}{lccccccc}
    \toprule
    \textbf{Method} & \textbf{MultiOff} & \textbf{Memotion 3 (A)} & \textbf{Memotion 3 (B)} & \textbf{FigMemes} & \textbf{MAMI (A)} & \textbf{MAMI (B)}\\
    \midrule

    TLC$_{norm}$ & 37.92 & \textbf{28.73} & \textbf{79.14} & 9.79  & 43.24 & \textbf{37.6} \\ 
    TLC$_{maj}$ & 52.78 & 18.88 & 77.6 & 9.08  & \textbf{52.92} & 29.89 \\
    TLC$_{rand}$ & \textbf{54.2} & 24.25 & 78.43 & \textbf{15.84} & 48.66 &  28.91 \\
     \bottomrule
    \end{tabular}
    \end{adjustbox}
    \caption{Classification results for the best-performing versions of TLC OOD and TLC$_{norm}$ (\textbf{in bold}) compared against each other.}
    \label{tlc_ood}
\end{table*}
As we are not dealing with probabilities but with a majority, this is reflected in our late fusion implementation. We use \emph{label vote} for both the template and its about section, combine all their labels, find the most common between the two, and keep that label as the final prediction for a given template. If we come across a template not featured in the training data, we back off to the most frequent label in the training split. For the datasets we explored, our implementation of late fusion was not a strong performer. This is intuitive because, as we have shown, using text representations is not as strong as image representations. Voting independently and then aggregating both modalities weakens image performance and is not as strong as other multimodal methods.

Additionally, in FigMemes, the authors tried many different models, for example, fine-tuning BERT \cite{devlin-etal-2019-bert}, which yielded a macro-averaged F1 of 32.62. TLC$_{Templates}$ is competitive with this model, but far cheaper. In Table 3 from their work, we see a great deal of variation, demonstrating the difficulty of the task.

\noindent
\subsection{Template Label Counter: Out of Distribution}
TLC's main drawback is its assumption that an input meme or image is employs a meme template. In order overcome this, we add an additional implementation of TLC that uses the same notion of \emph{templateness} as TSplit. That is, if the meme exceeds the maximum distance between template and examples, we declare it non-templatic. In this case, we back off to the most common label in the training data or randomly choose a label from the training data, which we refer to as TLC$_{maj}$ and TLC$_{rand}$, respectively. We compare this out of distribution implementation against making a prediction that assumes all memes are templatic, which we refer to as TLC$_{norm}$. This implementation also allows for more traditional hyperparameter search.
We ran extensive experiments with these alternative models, where we again examined performance with different encoders, number of neighbors, and combinations of features. Table \ref{tlc_ood} shows our results, where we show performance on the test set based on best performance on the validation data. If a validation split did not exist, we created one by sampling 20\% of the training data.

It is unsurprising that TLC$_{norm}$ is the strongest performer for both tasks of Memotion 3, as this dataset contains many templatic memes and TLC was already a strong performer here. FigMemes continues to be a difficult dataset with a complex labeling scheme which requires fine-tuning to be performant. All versions of TLC continue to struggle with MAMI, which we attribute to the low ratio of templatic memes in this dataset.

\subsection{Additional classification results}
\label{more_clf_results}
In this section, we provide additional results from our experiments that could not be put into the main text due to space limitations. Each table contains the results for a different type of modality or combination of modalities. Namely, we keep the modalities separate, we concatenate the embeddings, we fuse the embeddings via an element-wise product, or we normalize and average the embeddings. In each setting, we search over one to five neighbors as described in Section~\ref{discuss}. In the tables below, we present results organized by encoder, different CLIP models, namely \texttt{ViT-L/14@336px}, \texttt{ViT-B/32}, and \texttt{ViT-B/16},\footnote{\url{https://github.com/openai/CLIP/blob/main/clip/clip.py}} organized in each table in that order and also by the number of neighbors used for voting. The best configuration was chosen for Table~\ref{classification} in the main text. Note that TLC$_{About/OCR}$ is only present in cases where the modalities are not combined, because in the other cases text embeddings are combined with the template or the meme embeddings.

We find that \texttt{ViT-L/14@336px} usually results in the strongest performer, consistent with our findings with TSplit, but there are exceptions. In the case of MultiOff and Memotion 3 (B) and Memotion 3 (A), for example,  ViT-B/16 and ViT-B/32, respectively, were the best backbones for our method.  

It is only in cases where we consider both templates and examples (TLC$_{Templates+Instances}$) that neighbor voting improves the final prediction. We believe that this is an intuitive finding for two reasons: (\emph{i})~similar templates have unique, but broad semantics and convey concepts with related emotion charges, e.g.,~negative or positive sentiment. Therefore, templates that are similar would be nearby in the feature space. And (\emph{ii})~template instances are many, conveying a specific meaning, and can be noisy or combinations of distinct templates, as we demonstrated in Section~\ref{eda}. This crowded and noisy feature space results in neighbors that may be nearby markedly different templates.

We compute all evaluation measures using scikit-learn twice, where we set zero division equal to zero and to one, taking the max result between the two. We do this to avoid cases with zero in the denominator which can happen when precision (true positive + false positive) or recall (true positive + false negative) is equal to zero. This would make the f-score undefined. However, it is possible that this results in a sample-averaged F1 of $1.00$ if we make no predictions for a given label, artificially inflating the weighted- or macro-averaged F1 score. In this case, we report the lower value. 

In the earlier version of our work, we considered MEMEX in the main text, but removed it as it is no longer relevant to our analysis. However, we keep the results here for transparency.

\begin{table*} [h]
\centering
    \begin{adjustbox}{width=\textwidth}
    \begin{tabular}{lccccccc}
    \toprule
    \textbf{Method} & \textbf{MultiOff} & \textbf{Memotion 3 (A)} &\textbf{Memotion 3 (B)} & \textbf{FigMemes}  & \textbf{MEMEX} & \textbf{MAMI (A)} & \textbf{MAMI (B)}  \\
    \midrule
     \textit{ViT-L/14@336px}\\ 
TLC$_{About/OCR\quad1}$ & 44.43 & 27.12 & 76.58 & 21.14 & 46.02 & 60.43 & 35.19 \\ 
TLC$_{Templates\quad1}$ & 54.75 & 30.72 & 78.35 & 28.67 & 44.22 & 65.05 & 39.61 \\ 
TLC$_{Templates+Instances\quad1}$ & 58.58 & 34.59 & 76.91 & 27.99 & 43.01 & 67.44 & 38.92 \\ 
TLC$_{Templates+Instances\quad2}$ & 38.9 & 31.77 & 73.95 & 15.09 & 41.25 & 43.28 & 22.27 \\ 
TLC$_{Templates+Instances\quad3}$ & 43.56 & 32.15 & 74.49 & 18.54 & 41.11 & 51.51 & 26.05 \\ 
TLC$_{Templates+Instances\quad4}$ & 48.29 & 32.39 & 74.92 & 21.68 & 42.45 & 56.78 & 27.93 \\ 
TLC$_{Templates+Instances\quad5}$ & 45.66 & 33.0 & 75.65 & 23.05 & 43.87 & 60.89 & 32.73 \\ 
\midrule 
\textit{ViT-B/32}\\ 
TLC$_{About/OCR\quad1}$ & 48.29 & 27.06 & 77.6 & 20.86 & 43.56 & 58.7 & 33.8 \\ 
TLC$_{Templates\quad1}$ & 48.15 & 35.79 & 77.51 & 24.68 & 43.01 & 59.31 & 37.5 \\ 
TLC$_{Templates+Instances\quad1}$ & 48.33 & 28.68 & 76.74 & 28.4 & 43.64 & 63.68 & 38.35 \\ 
TLC$_{Templates+Instances\quad2}$ & 39.19 & 31.67 & 73.96 & 11.21 & 42.12 & 39.44 & 22.14 \\ 
TLC$_{Templates+Instances\quad3}$ & 40.94 & 32.63 & 75.13 & 15.44 & 42.12 & 45.71 & 23.87 \\ 
TLC$_{Templates+Instances\quad4}$ & 43.92 & 33.4 & 75.21 & 18.57 & 42.12 & 48.57 & 26.67 \\ 
TLC$_{Templates+Instances\quad5}$ & 43.2 & 34.1 & 75.81 & 21.04 & 42.12 & 52.3 & 29.74 \\ 
\midrule 
\textit{ViT-B/16}\\ 
TLC$_{About/OCR\quad1}$ & 51.83 & 35.4 & 76.2 & 20.29 & 46.25 & 59.04 & 35.44 \\ 
TLC$_{Templates\quad1}$ & 42.68 & 33.33 & 78.36 & 26.32 & 43.01 & 63.68 & 37.99 \\ 
TLC$_{Templates+Instances\quad1}$ & 51.32 & 36.42 & 78.13 & 26.87 & 43.71 & 63.31 & 37.34 \\ 
TLC$_{Templates+Instances\quad2}$ & 39.19 & 31.69 & 74.15 & 13.16 & 40.7 & 41.66 & 24.05 \\ 
TLC$_{Templates+Instances\quad3}$ & 39.99 & 51.58 & 74.56 & 15.95 & 42.25 & 48.43 & 27.21 \\ 
TLC$_{Templates+Instances\quad4}$ & 41.76 & 32.08 & 74.79 & 19.18 & 42.55 & 54.72 & 29.63 \\ 
TLC$_{Templates+Instances\quad5}$ & 42.3 & 32.45 & 75.11 & 22.27 & 42.55 & 56.98 & 32.23 \\ 
\midrule 

    \end{tabular}
    \end{adjustbox}
    \caption{TLC classification results where the text and the image modalities are kept separate. The results are organized by encoder and the number of neighbors used for voting.}
    \label{tlc4}
        \end{table*}

\begin{table*} [h]
\centering
    \begin{adjustbox}{width=\textwidth}
    \begin{tabular}{lccccccc}
    \toprule
    \textbf{Method} & \textbf{MultiOff} & \textbf{Memotion 3 (A)} &\textbf{Memotion 3 (B)} & \textbf{FigMemes}  & \textbf{MEMEX} & \textbf{MAMI (A)} & \textbf{MAMI (B)}  \\
    \midrule
     \textit{ViT-L/14@336px}\\ 
TLC$_{Templates\quad1}$ & 43.64 & 37.77 & 77.51 & 25.04 & 44.56 & 65.29 & 39.99 \\ 
TLC$_{Templates+Instances\quad1}$ & 45.29 & 28.5 & 78.6 & 23.81 & 41.07 & 69.09 & 38.07 \\ 
TLC$_{Templates+Instances\quad2}$ & 38.9 & 26.04 & 74.09 & 14.15 & 43.47 & 50.47 & 23.67 \\ 
TLC$_{Templates+Instances\quad3}$ & 43.2 & 27.07 & 75.57 & 18.24 & 42.88 & 54.58 & 27.92 \\ 
TLC$_{Templates+Instances\quad4}$ & 48.78 & 28.4 & 77.39 & 21.42 & 43.15 & 57.19 & 31.76 \\ 
TLC$_{Templates+Instances\quad5}$ & 48.74 & 29.18 & 77.36 & 23.53 & 43.33 & 60.4 & 34.74 \\ 
\midrule 
\textit{ViT-B/32}\\ 
TLC$_{Templates\quad1}$ & 52.56 & 27.62 & 76.35 & 26.59 & 44.84 & 60.42 & 37.87 \\ 
TLC$_{Templates+Instances\quad1}$ & 51.35 & 29.75 & 77.32 & 25.75 & 42.4 & 64.1 & 37.51 \\ 
TLC$_{Templates+Instances\quad2}$ & 41.61 & 33.02 & 74.95 & 12.99 & 40.7 & 46.44 & 23.86 \\ 
TLC$_{Templates+Instances\quad3}$ & 46.59 & 34.4 & 75.56 & 17.83 & 43.58 & 53.05 & 28.68 \\ 
TLC$_{Templates+Instances\quad4}$ & 52.24 & 34.45 & 76.25 & 19.59 & 42.38 & 57.71 & 31.84 \\ 
TLC$_{Templates+Instances\quad5}$ & 53.09 & 32.86 & 76.24 & 22.47 & 42.86 & 58.51 & 33.42 \\ 
\midrule 
\textit{ViT-B/16}\\ 
TLC$_{Templates\quad1}$ & 61.89 & 34.65 & 76.56 & 25.74 & 41.3 & 61.59 & 38.41 \\ 
TLC$_{Templates+Instances\quad1}$ & 53.98 & 35.76 & 78.65 & 23.65 & 43.77 & 62.33 & 37.09 \\ 
TLC$_{Templates+Instances\quad2}$ & 47.01 & 32.6 & 74.75 & 13.16 & 40.7 & 48.06 & 24.14 \\ 
TLC$_{Templates+Instances\quad3}$ & 49.07 & 33.44 & 76.06 & 18.48 & 43.37 & 53.84 & 29.29 \\ 
TLC$_{Templates+Instances\quad4}$ & 49.06 & 27.28 & 76.89 & 19.54 & 42.12 & 57.64 & 31.2 \\ 
TLC$_{Templates+Instances\quad5}$ & 50.83 & 35.28 & 77.62 & 20.8 & 43.0 & 59.78 & 33.17 \\ 
\midrule 

    \end{tabular}
    \end{adjustbox}
    \caption{TLC classification results where the text and the image modalities are concatenated. The results are organized by encoder and the number of neighbors used for voting.}
    \label{tlc3}
        \end{table*} 

\begin{table*} [h]
\centering
    \begin{adjustbox}{width=\textwidth}
    \begin{tabular}{lccccccc}
    \toprule
    \textbf{Method} & \textbf{MultiOff} & \textbf{Memotion 3 (A)} &\textbf{Memotion 3 (B)} & \textbf{FigMemes}  & \textbf{MEMEX} & \textbf{MAMI (A)} & \textbf{MAMI (B)}  \\
    \midrule
     \textit{ViT-L/14@336px}\\ 
TLC$_{Templates\quad1}$ & 43.13 & 30.56 & 79.89 & 19.44 & 44.99 & 54.06 & 33.43 \\ 
TLC$_{Templates+Instances\quad1}$ & 51.26 & 36.48 & 80.17 & 18.76 & 48.14 & 57.99 & 35.4 \\ 
TLC$_{Templates+Instances\quad2}$ & 43.92 & 32.63 & 74.78 & 25.76 & 41.05 & 39.91 & 22.27 \\ 
TLC$_{Templates+Instances\quad3}$ & 38.71 & 33.2 & 75.63 & 13.02 & 40.9 & 45.13 & 23.82 \\ 
TLC$_{Templates+Instances\quad4}$ & 45.46 & 33.65 & 77.22 & 13.14 & 40.86 & 48.21 & 26.38 \\ 
TLC$_{Templates+Instances\quad5}$ & 45.32 & 35.93 & 77.06 & 15.24 & 41.1 & 48.2 & 27.89 \\ 
\midrule 
\textit{ViT-B/32}\\ 
TLC$_{Templates\quad1}$ & 49.68 & 26.88 & 78.62 & 21.37 & 42.97 & 60.68 & 31.73 \\ 
TLC$_{Templates+Instances\quad1}$ & 52.83 & 27.36 & 78.05 & 18.46 & 44.6 & 53.11 & 32.84 \\ 
TLC$_{Templates+Instances\quad2}$ & 41.57 & 32.26 & 74.59 & 41.83 & 41.05 & 38.81 & 20.13 \\ 
TLC$_{Templates+Instances\quad3}$ & 42.29 & 29.95 & 75.32 & 27.24 & 41.05 & 42.97 & 22.96 \\ 
TLC$_{Templates+Instances\quad4}$ & 45.85 & 30.74 & 75.04 & 28.97 & 41.5 & 45.77 & 25.36 \\ 
TLC$_{Templates+Instances\quad5}$ & 45.25 & 33.82 & 74.82 & 14.72 & 43.66 & 46.01 & 26.24 \\ 
\midrule 
\textit{ViT-B/16}\\ 
TLC$_{Templates\quad1}$ & 49.29 & 29.56 & 77.35 & 19.57 & 43.47 & 56.3 & 32.81 \\ 
TLC$_{Templates+Instances\quad1}$ & 50.09 & 29.52 & 80.49 & 19.64 & 43.49 & 54.18 & 33.51 \\ 
TLC$_{Templates+Instances\quad2}$ & 44.65 & 25.71 & 74.48 & 41.26 & 40.7 & 36.84 & 20.67 \\ 
TLC$_{Templates+Instances\quad3}$ & 48.14 & 32.56 & 75.89 & 9.84 & 40.7 & 41.12 & 22.95 \\ 
TLC$_{Templates+Instances\quad4}$ & 50.02 & 34.14 & 76.36 & 12.0 & 41.52 & 46.16 & 24.85 \\ 
TLC$_{Templates+Instances\quad5}$ & 47.44 & 33.78 & 75.96 & 14.44 & 42.12 & 47.78 & 26.11 \\ 
\midrule 

    \end{tabular}
    \end{adjustbox}
    \caption{TLC classification results where the text and the image modalities are fused via the Hadamard product. The results are organized by encoder and the number of neighbors used for voting.}
    \label{tlc2}
        \end{table*}

\begin{table*} [h]
\centering
    \begin{adjustbox}{width=\textwidth}
    \begin{tabular}{lccccccc}
    \toprule
    \textbf{Method} & \textbf{MultiOff} & \textbf{Memotion 3 (A)} &\textbf{Memotion 3 (B)} & \textbf{FigMemes}  & \textbf{MEMEX} & \textbf{MAMI (A)} & \textbf{MAMI (B)}  \\
    \midrule
     \textit{ViT-L/14@336px}\\ 
TLC$_{Templates\quad1}$ & 48.72 & 27.81 & 76.88 & 29.8 & 44.4 & 62.7 & 37.77 \\ 
TLC$_{Templates+Instances\quad1}$ & 52.89 & 37.04 & 77.58 & 25.5 & 46.01 & 63.01 & 36.57 \\ 
TLC$_{Templates+Instances\quad2}$ & 46.48 & 33.06 & 75.84 & 16.43 & 44.21 & 51.55 & 27.12 \\ 
TLC$_{Templates+Instances\quad3}$ & 40.96 & 26.11 & 75.96 & 18.78 & 45.2 & 58.25 & 30.93 \\ 
TLC$_{Templates+Instances\quad4}$ & 46.07 & 26.63 & 75.5 & 22.08 & 45.52 & 61.23 & 32.41 \\ 
TLC$_{Templates+Instances\quad5}$ & 47.9 & 25.99 & 77.13 & 23.45 & 45.36 & 62.79 & 33.83 \\ 
\midrule 
\textit{ViT-B/32}\\ 
TLC$_{Templates\quad1}$ & 57.09 & 33.55 & 78.04 & 25.07 & 42.45 & 60.65 & 35.95 \\ 
TLC$_{Templates+Instances\quad1}$ & 49.07 & 35.22 & 77.75 & 23.36 & 43.99 & 63.21 & 36.96 \\ 
TLC$_{Templates+Instances\quad2}$ & 43.2 & 32.18 & 74.07 & 13.71 & 41.1 & 50.66 & 28.94 \\ 
TLC$_{Templates+Instances\quad3}$ & 42.12 & 32.55 & 75.27 & 16.76 & 42.15 & 56.41 & 28.41 \\ 
TLC$_{Templates+Instances\quad4}$ & 41.71 & 25.7 & 75.96 & 19.2 & 42.31 & 59.1 & 32.54 \\ 
TLC$_{Templates+Instances\quad5}$ & 44.02 & 25.37 & 76.46 & 20.08 & 42.93 & 63.12 & 33.12 \\ 
\midrule 
\textit{ViT-B/16}\\ 
TLC$_{Templates\quad1}$ & 47.54 & 34.57 & 75.15 & 24.39 & 42.6 & 64.43 & 38.72 \\ 
TLC$_{Templates+Instances\quad1}$ & 47.7 & 27.46 & 78.59 & 24.04 & 42.82 & 61.49 & 35.41 \\ 
TLC$_{Templates+Instances\quad2}$ & 50.02 & 33.25 & 74.88 & 13.41 & 43.84 & 51.08 & 24.79 \\ 
TLC$_{Templates+Instances\quad3}$ & 44.36 & 32.12 & 76.03 & 18.93 & 43.97 & 56.67 & 28.13 \\ 
TLC$_{Templates+Instances\quad4}$ & 49.19 & 25.41 & 76.11 & 20.61 & 44.34 & 58.51 & 30.04 \\ 
TLC$_{Templates+Instances\quad5}$ & 49.22 & 33.77 & 77.29 & 22.0 & 44.34 & 59.34 & 32.06 \\ 
\midrule 

    \end{tabular}
    \end{adjustbox}
    \caption{TLC classification results where the text and the image modalities are normalized and averaged. The results are organized by encoder and the number of neighbors used for voting.}
    \label{tlc1}
        \end{table*}

\clearpage
\newpage
\noindent

\begin{table} [h!]
\centering

    \resizebox{.5\textwidth}{!}{%
    \begin{tabular}{lccccccc}
    \toprule
    \textbf{Method} & \textbf{MultiOff} & \textbf{Memotion 3 (A)} &\textbf{Memotion 3 (B)} & \textbf{FigMemes}  & \textbf{MAMI (A)} & \textbf{MAMI (B)}  \\
    \midrule
    TLC$_{best}$ & 61.89 & 37.77 & 80.49 & 29.8 & 70.0 & 40.21 \\
    \midrule
TLC$_{TSplit\quad1}$ & 37.12 & 22.8 & 70.93 & 5.96 & 29.5 & 18.08  \\ 
TLC$_{TSplit\quad2}$ & 49.27 & 29.7 & 72.84 & 15.86 & 63.82 & 28.54  \\
TLC$_{TSplit\quad3}$ & 45.05 & \textbf{32.08} & 76.75 & 21.46 & 69.12 & 32.99  \\ 
TLC$_{TSplit\quad4}$ & \textbf{51.81} & 28.8 & 74.78 & 28.67 & 75.2 & \textbf{37.53}  \\ 
TLC$_{TSplit\quad5}$ & 48.7 & 31.62 & \textbf{77.39} & \textbf{29.55} & \textbf{75.73} & 37.35  \\

\end{tabular}
}
    \centering
    \caption{TLC classification results after TSplit preprocessing (the highest result is in \textbf{bold}).}
    \label{tab_tsplit_tlc}
    
        \end{table} 
\subsection{TSplit then TLC}
\label{tsplitthentlc}
In this Appendix section, we present TLC results after preprocessing via TSplit (see Table \ref{tab_tsplit_tlc}). We searched over one to five neighbors, using only the base template. We took a tolerant view of what constitutes a template instance, using TSplit$_{max}$.

TLC assumes all memes are templatic and relies on the template signal, assuming that the most common label for a given template in the training data is the correct label for an instance of that template in the test data. However, TSplit ensures that templates do not overlap between dataset splits, effectively removing TLC's predictive power. Under these conditions, TLC is far more likely to assign the wrong template and therefore likely the wrong label. 

When a dataset contains many templatic memes, such as Multioff or Memotion 3, we see that TLC's performance drops markedly, even if we consider additional base templates (additional neighbors). However, when a dataset does not contain as many templatic memes, MAMI, or is very difficult, FigMemes, the effect on performance is not as dramatic, and voting with multiple templates allows us to achieve similar if not improved performance relative to the best performing version of TLC from Table \ref{classification}. In the case of MAMI, this is because TLC alone was already assigning templates erroneously most of the time. For FigMemes, this is due to the difficulty of the task, where assigning a fixed interpretation for a given template is too simple an approach for a nuanced labeling system (see Section \ref{dataset_fino}) and also a nuanced means of communication like memes.

We searched over one to five neighbors and include the best TLC results from Table \ref{classification} for reference. Performance suffers when a dataset is highly templatic, while it is not as affected or even improves when the dataset does not contain as many templatic memes. \texttt{ViT-L/14@336px} was used as the CLIP encoder.

\begin{figure}[h] 
    \centering
    \includegraphics[width=.45\textwidth]{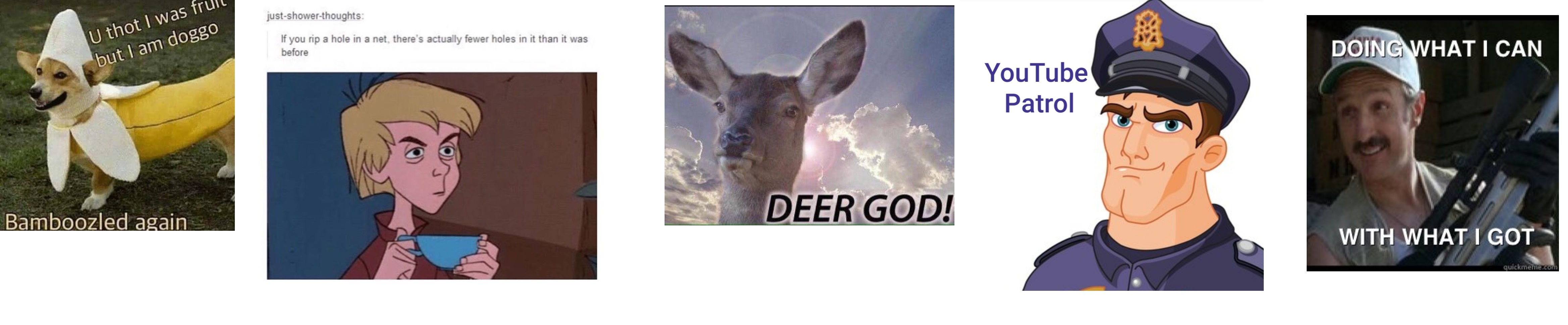}
    \caption{Examples of non-template images found in FigMemes.}
    \label{non_templatic}
\end{figure}

\subsection{Non-Templatic Memes}
\label{notemplates}

In this Appendix section, we provide examples of images/memes which we consider to be non-templatic (see Figure \ref{non_templatic}). The first and third examples are a visual joke and pun respectively. The first does make reference to the \emph{Doggo}\footnote{\url{https://knowyourmeme.com/memes/doggo}} and language from the \emph{Cheezburger}\footnote{\url{https://knowyourmeme.com/memes/sites/cheezburger}} 
templates.
The second is a still from a .gif that references the film \textit{The Sword in the Stone}\footnote{\url{https://en.wikipedia.org/wiki/The_Sword_in_the_Stone}} and can actually be found on KYM.

This is a bit of an edgecase, but we believe this communicates the idea of confusion or realization triggered by the text above the image, which is interpretable without knowledge of a template. The fourth example is arguably not a meme and we actually are not sure of the interpretation without additional context. A possible reading, given the domain of FigMemes, is a criticism of users who comment on YouTube, forcing their views of social norms on the politically incorrect, but this a forced interpretation. 
The fifth example is a still from the movie \textit{Tremors II: Aftershock}, where the image is the correct character but the text is quoted anachronistically from another part of the film.
\begin{figure*}[t]%
    \centering
    \includegraphics[width=.85\textwidth]{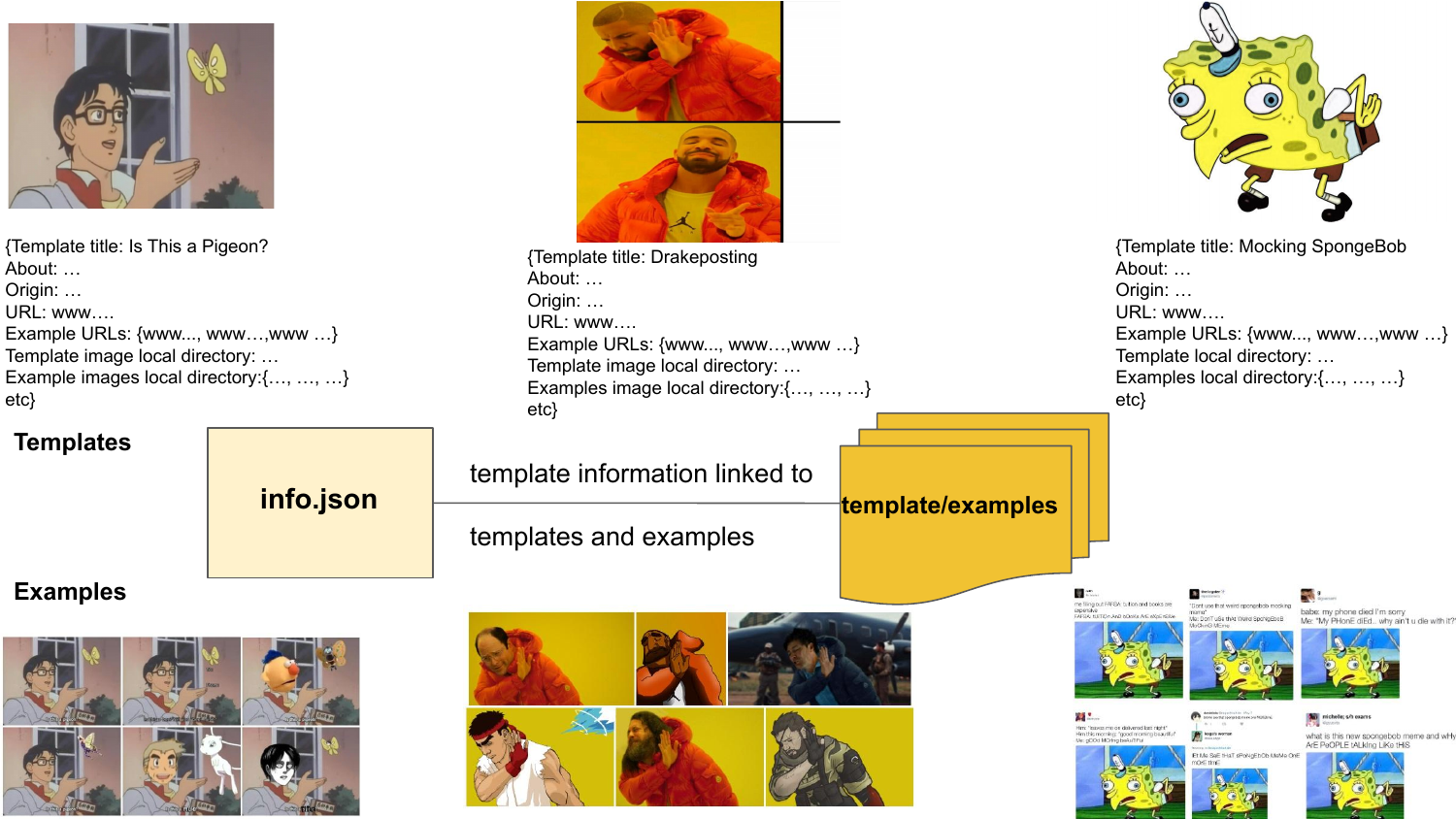}
    \caption{The KYMKB records all textual information about a meme in a .json file, including the text found on KYM, the URLs used in the scraping process, and local locations of all template and example images in the knowledge base.}
    \label{kymkb}
\end{figure*}
All information relevant to the scraping process is preserved in a .json file, linking templates to their examples (see Figure \ref{kymkb}).
\label{scrape}
\subsection{Scraping details}
We use WM for all scraping and use the most recent snapshots available. We first visit the main page of KYM on WM with Selenium\footnote{\url{https://selenium-python.readthedocs.io/}} and autoscroll through the entire meme table. We collect and write .htmls files for each cell in the template table. We then process the .html files, parsing each meme template and collecting the text appearing in the table with Beautiful Soup.\footnote{\url{https://beautiful-soup-4.readthedocs.io/en/latest/}} We collect all urls seen and write them to a .json file. This ensures that we do not revisit seen pages if the scraper is interrupted and it very likely will be. To collect template instances (examples), we look at each image url and determine if WM has a snapshot available and collect the image and write to disk if it exists. Again, we record all visited urls to avoid revisiting and associate each image with its parent template see Figure \ref{kymkb}.

WM's snapshots of the Internet are incomplete, making it impossible to completely capture KYM via WM; of the roughly 8,400 confirmed entries at the time of writing, we were only able to scrape 5,220. However, we are passionate about memes and we are devoted to making the KYMKB as complete as possible. We therefore release all our scraping code such that other members of the community can contribute to and improve meme analysis grounded in templates.

\subsection{More template-meme analysis}
In this Appendix section, we showcase additional examples of how the KYMKB can be easily used with simple, well-known algorithms, such as nearest neighbor indexing and $k$-means clustering to gain insight into a meme dataset. Specifically, we can use retrieval to examine how well templates in our knowledge base map onto memes in a dataset. Often we find that the memes in a dataset are simply the base templates or examples already contained in the knowledge base. We argue that examination of cluster centroids yields insight into which templates best reflect the type of memes in a dataset. For example, FigMemes was collected from 4chan /pol/, and by investigating cluster centroids we unintentionally arrived at the /pol/ template. We emphasize that we did nothing but consider the template closest to a centroid and arrive at a template we ourselves were previously unaware of. Details can be found below. 
\label{more_eda}
\begin{figure*}[t]%
    \centering
    \includegraphics[width=.85\textwidth]{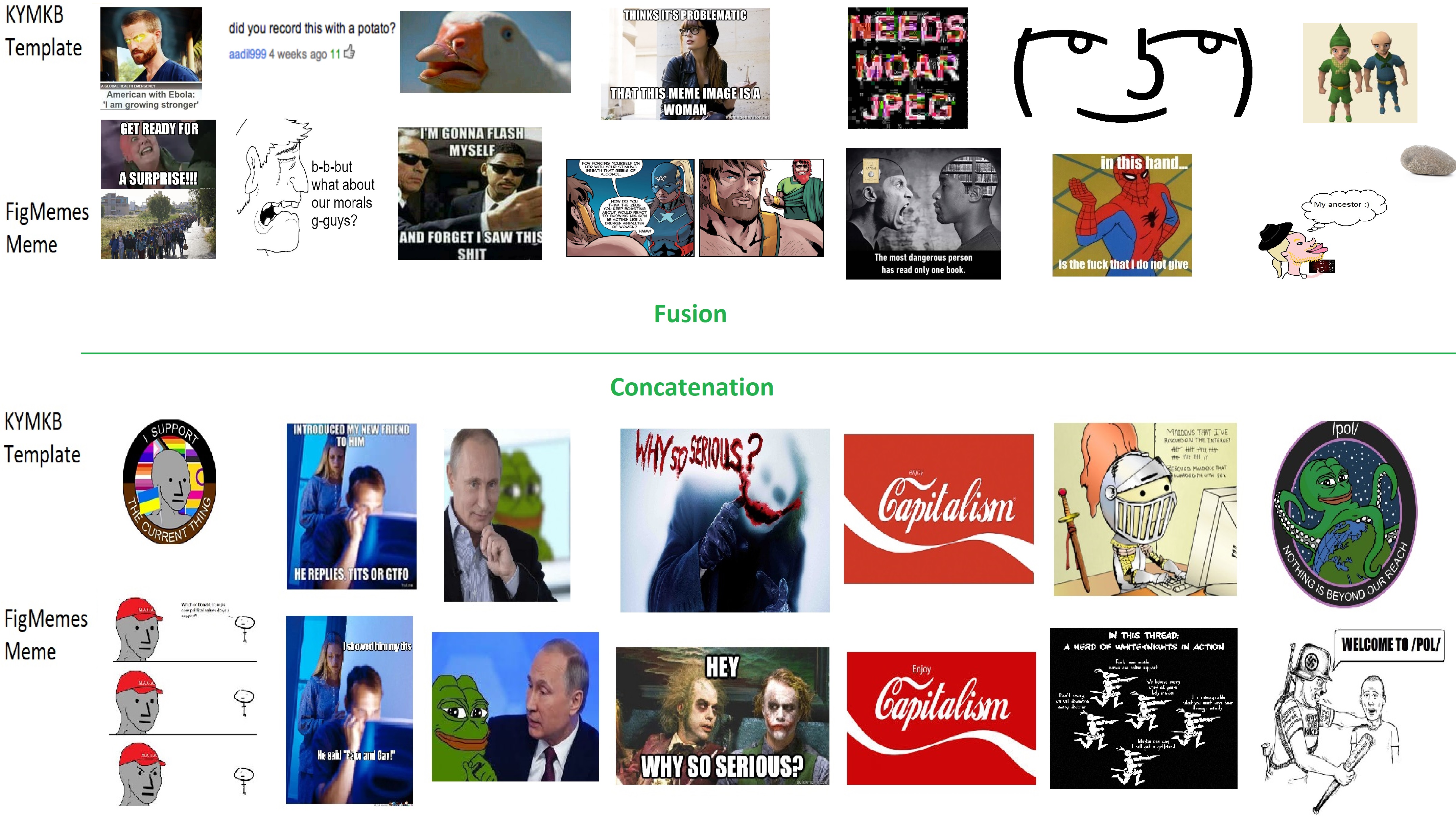}
    \caption{KYMKB templates matched via similarity search to FigMemes images.}
    \label{retrieve appendix}
\end{figure*}
\subparagraph{Retrieval} Here we provide further examples and details regarding the retrieval-based examination of the KYMKB from the main text, Section~\ref{eda}. After querying the 500 closest neighbors, we then randomly select $k$ pairs, where $k$ is equal to the number of labels in a given dataset. The pairs, as in the main text, are composed of the template and its nearest neighbor in the dataset. For conciseness, we only consider FigMemes here as it is a difficult dataset with the most labels.
 
Figure \ref{retrieve appendix} shows a sample of our findings. Combining embeddings via fusion or normalizing and averaging the vectors results in matches where the relation between a template and a meme is nuanced or nonexistent.

We again find that either only considering the image modality or concatenating the image and the text representations results in the strongest signal, and indeed, using this configuration for retrieval makes it difficult to appear as though we are not cherry-picking. We clearly match either a base template to a meme or a base template to an obvious instance of that template. In cases when it is not so obvious, we match text or characters, such as \emph{Why So Serious} or the \emph{Joker},\footnote{Note that this text and character have taken on lives on their own in meme culture. \url{https://knowyourmeme.com/memes/why-so-serious}} or concepts that exist in only meme or Internet culture. For example, consider the first column under the concatenation setting in Figure~\ref{retrieve appendix}. We observe the character of Wojak in the \emph{I Support the Current Thing} meme template,\footnote{\url{https://knowyourmeme.com/memes/npc-wojak}} \footnote{\url{https://knowyourmeme.com/memes/i-support-the-current-thing}} a template that criticizes social media users for being a simpleton or lacking critical thinking skills. We match this template to a meme criticizing Trump supporters for the same faults, despite drastically different appearances. In the sixth column, we match the template of White Knight to an image that derides \emph{White Knighting}.\footnote{\url{https://knowyourmeme.com/memes/white-knight}} This template and its entry in the KYMKB provide sufficient background to interpret the FigMemes image, which is arguably not even a meme. Finally, in the seventh column, we match the template of /pol/ to a meme obviously about the 4chan board.\footnote{\url{https://knowyourmeme.com/memes/sites/pol}} We share this information not to explain memes, but to demonstrate the ease and the power of using the KYMKB to retrieve information about not only memes, but also images related to Internet culture. If one is not familiar with these concepts, it is difficult to even know what to search for; however, this is different with KYMKB.

\begin{figure*}[t]%
    \centering
    \includegraphics[width=.85\textwidth]{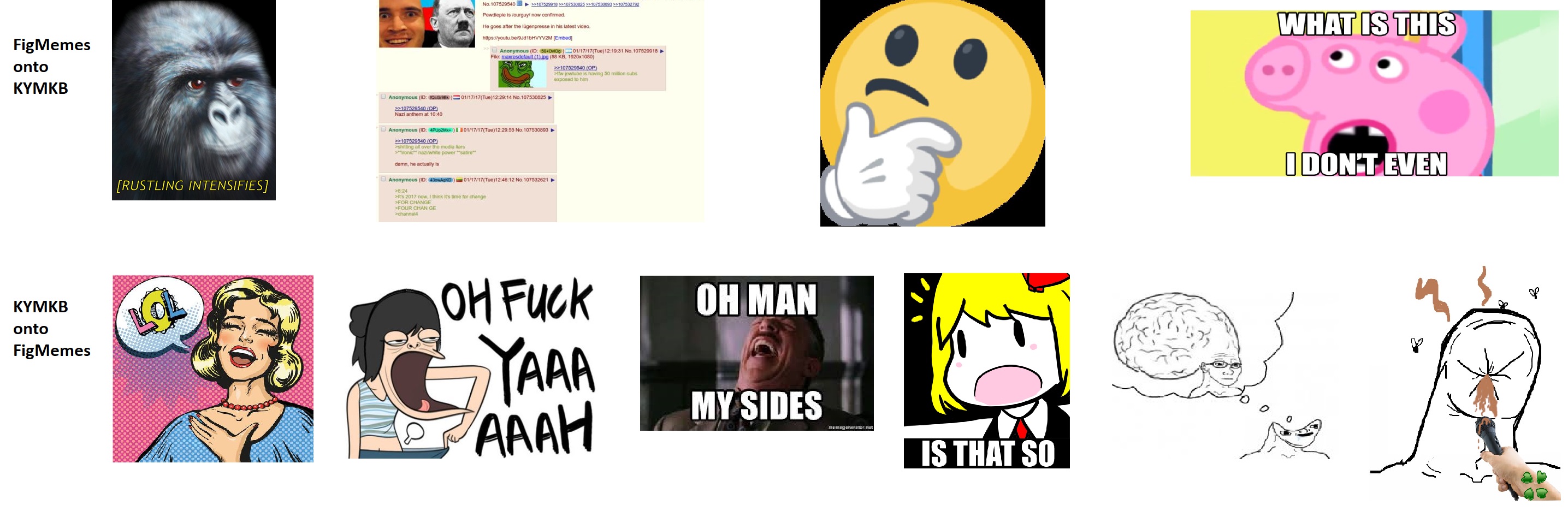}
    \caption{In the first row, we show the templates closest to seven KMeans centroids fit on the FigMemes, while in the second row, we show FigMeme images closest to seven centroids derived from KYMKB. We combine the text and the image representations by normalizing and averaging the two modalities. This results in multiple centroids close to the same meme/template.}
    \label{cluster1}
\end{figure*}
\begin{figure*}[ht]%
    \centering
    \includegraphics[width=.85\textwidth]{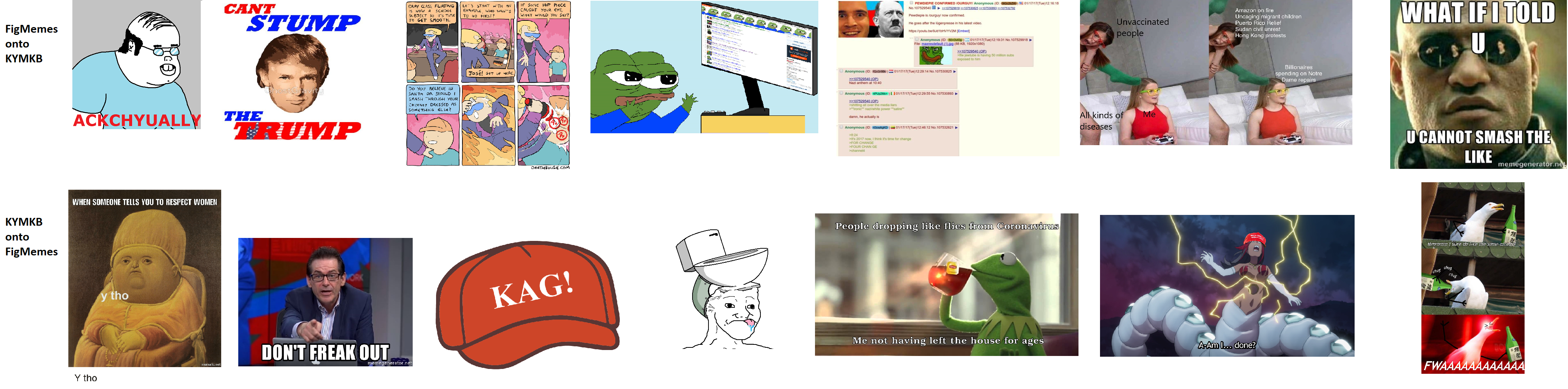}
    \caption{In the first row, we show the templates closest to seven KMeans centroids fit on the FigMemes, while in the second row, we show FigMeme images closest to seven centroids derived from the KYMKB. We only use the image modality, which results in seven distinct images.}
    \label{cluster2}
\end{figure*}

\subparagraph{Clustering} In order to investigate the saliency of templatic memes in the context of meme datasets, we conduct distance-based clustering using KMeans where we fit the algorithm on both the KYMKB, with or without examples, and on the dataset in question, encoding all memes using CLIP. We then manually examine the closest meme or template to each centroid, respectively. We set $k$ to be equal to the number of labels in each dataset (see Table~\ref{datasum}). Here, for conciseness, we consider only templates and FigMemes, as we consider it a difficult dataset and it has the most labels/.

Figures \ref{cluster1} and \ref{cluster2} show a sample of our results. If we attempt to combine the image and the text embeddings, either via fusion or normalization and averaging, we find that this often results in repeated images, that is, a meme or a template is close to multiple centroids. However, if we concatenate the embeddings or only use the images representation, we find that we are left with centroids that point to $k$ distinct image files, where $k$ is again equal to the number of labels in a given dataset: seven in the case of FigMemes.

Naturally, when we consider centroids fit on KYMKB, their closest meme in FigMemes reflects the nature of that dataset. These memes express sexist or politically charged, but still toxic rhetoric, which 4chan /pol/ is known for. Somewhat surprisingly, when we determine the centroids from the dataset and query the closest template in the KYMKB, we again see the nature of the dataset reflected, where we had expected to be met with potentially political, but not toxic templates. The resulting image files express salient traits of derision, sexism, or conservative political beliefs. Interestingly, if we combine modalities or only consider image representations, one meme centroid is closest to the same template in both cases, that is the \emph{Is He /Our Guy/?} template.\footnote{\url{https://knowyourmeme.com/memes/is-he-our-guy}} This 4chan-specific template is used to confirm whether a celebrity shares similar beliefs as the ``politically incorrect'' community, e.g.,~supporting Nazism. It is surprising that an examination of centroids in this way provides such a succinct summary of the domain of the dataset.

\subparagraph{Annotation}
\label{agreement}
In this Appendix section, we provide additional details on the annotation conducted in Section \ref{eda}. Our annotators were two colleagues of ours that are knowledgeable about memes. They were warned that the template-meme pairs might be offensive.

\subsection{Meme ``edge cases''}
\label{edging}
\begin{figure*}[t]%
    \centering
    \includegraphics[width=.85\textwidth]{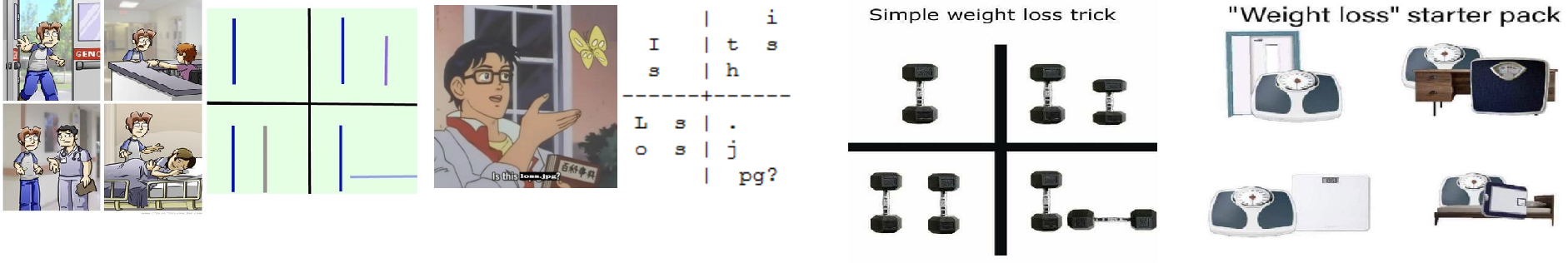}
    \caption{The first image is the original template of \emph{Loss}, while the other three images are \emph{Loss} instances, all of which are visual puns that cannot be understood without knowing the original template. The second image is another intertextual meme where \emph{Loss} and \emph{Is This a Pigeon?} have been amalgamated.}
    \label{loss}
\end{figure*}

Below, we provide a discussion and background on examples of meme templates contained in the KYMKB that defy the narrow scope of memes being static images. The templates we discuss are by no means exhaustive and we provide this section purely as additional motivation for our argument that the AI community must not limit itself simply to static images. 

One of the oldest templates is \emph{Rickroll},\footnote{\url{https://knowyourmeme.com/memes/rickroll}} which can involve posting an image of Rick Astley from the Never Going to Give You Up music video,\footnote{\url{https://www.youtube.com/watch?v=dQw4w9WgXcQ}} but more frequently an instance of this template is a bait-and-switch prank where posters trick others into viewing the music video. This has since evolved where the prank is now to trick others into stating the title of the song.\footnote{\url{https://knowyourmeme.com/photos/1901413-rickroll}} We would argue this is an intertextual meme instance referencing the Rickroll template.

\emph{Loss}\footnote{\url{https://knowyourmeme.com/memes/loss}} is another famous template where an instance is an action, not an image. The template is a reference to the Ctrl+Alt+Del Comic\footnote{\url{https://cad-comic.com/}} gaming webcomic, which made an uncharacteristically serious update about a miscarriage. The idea that this webcomic could approach such a serious topic amused many social media users, and they began mocking the strip by posting references to the panel as a joke, bringing it to its meme status. The strip was referenced so ubiquitously that the positions of the characters in the strip, that is, one vertical line, two vertical lines of different heights, two vertical lines of the same height, and one vertical and one horizontal line became an instance of this template. The phrase \emph{Is this Loss?} became a meme by itself, as users wondered whether certain posts or memes were instances of the \emph{Loss} template (see Figure~\ref{loss}).

Instances of the \emph{Planking}\footnote{\url{https://knowyourmeme.com/memes/planking}} template is again a behavior where a person lies flat on their stomach with their arms to their sides in an unusual place, has their photo taken, and uploads this for the amusement of others.

Another tricky template is that of \emph{Thinking Face Emoji}.\footnote{\url{https://knowyourmeme.com/memes/thinking-face-emoji}} An instance of this template would be ironically or sarcastically posting a thinking face emoji. However, this could be simply using the Unicode "U+1F914" or posting a picture of the emoticon for extra emphasis.

A recent example of a meme that is not an image is the \emph{OOF / Roblox Death Sound} template.\footnote{\url{https://knowyourmeme.com/memes/oof-roblox-death-sound}} An instance of this template is featuring or remixing the audio clip in videos or music, referencing an amusing sound effect from the popular MMORPG Roblox.\footnote{\url{https://www.roblox.com/}} Players of this game found the audio clip so amusing that it is referenced to suggest humorously express empathy for another's misfortune and shared experience.

\subsection{Datasets and previous work details}
\label{dataset_fino}
\begin{table*} [t]
\centering
    \begin{adjustbox}{width=\textwidth}
    \begin{tabular}{llcccccc}
    \toprule
    \textbf{Dataset} & \textbf{Task} & \textbf{Number of Labels} & \textbf{Size} & \textbf{Multilabel?} & \textbf{Multilingual?} & \textbf{Evaluation Measure} \\
    \midrule
     FigMemes & Figurative Language & 7 & 5141 &  Yes & No & Macro-F1 \\
     MultiOff & Offensive Language & 2 & 743 & No & No & Macro-F1 \\
     MAMI Task A & Misogyny Detection & 2 & 11k & No & No & Macro-F1\\
     MAMI Task B & Types of Misogyny & 4 & 11k & Yes & No & Weighted-F1\\
     Memotion 3 Task A & Sentiment Analysis & 3 & 10k & No & Yes & Weighted-F1  \\
     Memotion 3 Task B & Types of Emotion & 4 & 10k &  Yes & Yes & Weighted-F1  \\
     \bottomrule
    \end{tabular}
    \end{adjustbox}
    \caption{Summary of the previous work we examine.}
    \label{datasum}

\end{table*}

\begin{table*} [t]
\centering
    \begin{adjustbox}{width=\textwidth}
    \begin{tabular}{lcccccc}
    \toprule
    \textbf{Dataset} & \textbf{Text Model} & \textbf{Vision Model} & \textbf{Multimodal Model} & \textbf{Agreement} \\
    \midrule
     FigMemes & DeBERTa & CLIP & CLIP &  0.42 \\
     MultiOff & GloVe + CNN & VGG 16 & Stacked LSTM + GloVe (text) / VGG 16 (image) (early fusion) & 0.2 -  0.3 to 0.4 - 0.5 \\
     MAMI Task A & NA & NA & Ensemble of XGBoost + CLIP + UNITER + BERT  & 0.5767 \\
     MAMI Task B & NA & NA & Ensemble of XGBoost + CLIP + UNITER + BERT  & 0.3373 \\
     Memotion 3 Task A & NA & NA & Hinglish BERT (text) / ViT (image) & Majority vote  \\
     Memotion 3 Task B & NA & NA & Hinglish BERT (text) / ViT (image) &  Majority vote  \\
     \bottomrule
    \end{tabular}
    \end{adjustbox}
    \caption{Continued summary of the previous work we examine. In the case of multimodal models, we provide them as \textit{text model} / \textit{vision model}. For agreement, we provide multiple scores to indicate that the researchers consulted their annotators, which led to an increase in agreement.}
    \label{modelsum}

\end{table*}
In this Appendix section, we provide additional information about the datasets we examined, such as their respective label inventories, distributions, and reported inter-annotator agreement scores. We also provide information on the models reported in the respective work. We do this for ease of reference and simply reproduce reported information where possible. When this information is not available, we report the information we are able to access.

MultiOff is a binary classification task, offensive (40\%) vs. not offensive (60\%), composed of memes related to the 2016 US Presidential Election. They report two Fleiss Kappas both before and after getting feedback from their annotators. The first is between 0.2 and 0.3 (fair agreement), while the other, after feedback, is between 0.4 and 0.5 (moderate agreement). Their text-based model is a combination of GloVe embeddings \cite{pennington-etal-2014-glove} and a CNN, while their image-based model was VGG 16 \cite{vgg16} pretrained on ImageNet \cite{imagenet}, and their multimodal method consistent of a stacked LSTM (text) and VGG 16 (image) combined via early fusion.

Memotion 3 is composed of two multilabel tasks (A and B). The test split is not publicly available, so we consider only the training and validation split. Task A is sentiment analysis for memes, where labels can be very positive (5\%), positive (26\%), neutral (42\%), negative (23\%), or very negative (5\%). Task B considers memes with humorous (39\%), sarcastic (37\%), offensive (19\%), and motivational (5\%) messages. They do not report inter-annotator agreement scores, settling disagreements via majority vote. This work reports only multimodal results, fusing and fine-tuning a BERT-based Hindi and English model \cite{hinglishnlp} for textual features and the ViT model \cite{dosovitskiy2021image} for image features. Note that their test split is not public at time of writing.

FigMemes is a multilabel task of determining the type of figurative language used in a meme. There are seven labels, composed of Allusion (17\%), Exaggeration (19\%), Irony (20\%), Anthropomorphism (9\%), Metaphor (20\%), Contrast (10\%), and None (30\%) (see the work for more information). They report a Fleiss Kappa of 0.42, indicating moderate agreement. The authors fine-tuned DeBERTa \cite{he2021deberta} for their text classifier and used various CLIP fine-tuning strategies for their image and multimodal experiments.

Task A in MAMI looks at whether memes are misogynous or not. The task has a balanced binary label distribution and the authors report a Fleiss-k of 0.5767. Task B examines different types of misogyny expressed in a meme. There are four labels, Shaming (17\%), Stereotype (38\%), Objectification (31\%), Violence (13\%), and the remaining do not express misogyny. The authors report a Fleiss-k of 0.3373, showing that is too is quite a difficult task. The best performing methods on this dataset are reported in \citet{zhang-wang-2022-srcb}, which involved multimodal fine-tuning and ensembling of XGBoost \cite{xgboost}, CLIP, UNITER \cite{chen2020uniter1}, and BERT.


See Tables \ref{datasum} and \ref{modelsum} above for a summary of this information.

\subsection{Prompting with LLMs}
\label{prompting}
We now explore if the KYMKB can also be used to aid a vision language model by grounding the model in a meme-specific context. We experiment with LLaVA \cite{liu2023improved} which employs LLaMA \cite{touvron2023llama} and CLIP to combine both the textual and visual input modalities.

We mainly conduct few-shot in-context-learning (ICL) experiments using one completion and three examples for text-only modality inputs, as the model has not been trained to handle multiple input images. We investigate the following scenarios, with and without RAG-style \cite{rag} prompting:

\begin{itemize}

    \item Which input modality is the most useful? Text, image, or both?
    \item Does providing a description of the labels in the prompt help? (i.e the meaning of "anthropomorphic" in FigMemes)
    \item Does performance increase if we retrieve the nearest meme template title and add it to the prompt? (i.e \emph{Drakeposting},\footnote{\url{https://knowyourmeme.com/memes/drakeposting}} \emph{Is This a Pigeon?}, etc.)
    \item Does LLM performance increase when we consider all examples of a template and not just the base template as candidates for retrieval?
    \item Does it help to discard retrieved meme information from the KYMKB if a given input meme is too different (not considered in distribution) from its closest entry in the KYMKB?
    \item Does the inclusion of the retrieved template \emph{about} section in the prompt improve LLM performance? 

\end{itemize}

We perform an extensive ablations to answer the above questions and the main results are shown in Table \ref{overview}. Our overall setup is illustrated in Figure~\ref{llava_kymkb}. 
We highlight the following results by answering the points raised above when grounding LLaVA in the KYMKB:

\begin{figure*}[t]%
    \centering
    \includegraphics[width=.85\textwidth]{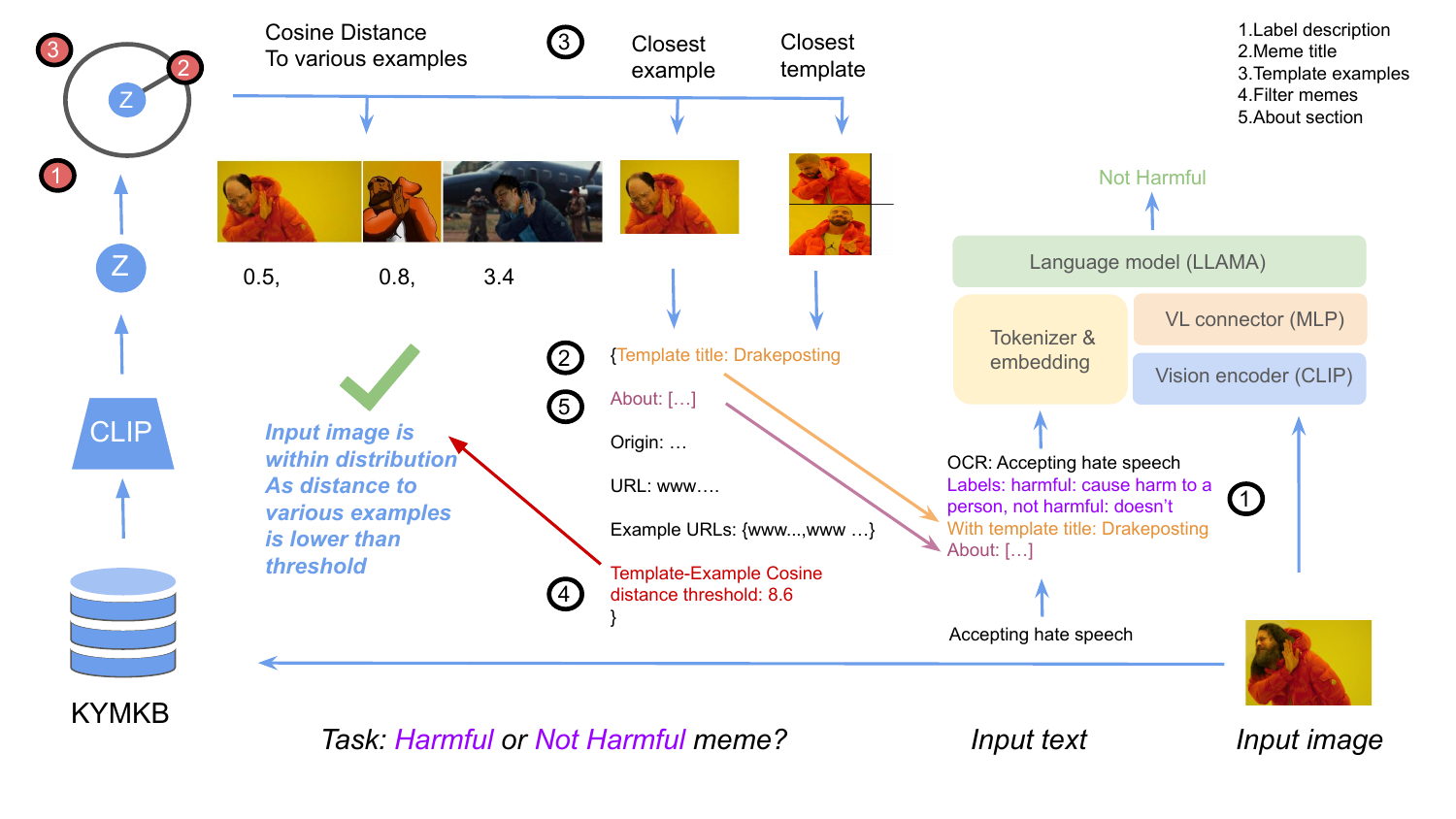}
    \caption{The complete setup of using LLAVA with KYMKB. All experiments are with multiple examples as illustrated in Figure ~\ref{tab:prompt-setup}. Initially, both the OCR text and input meme are passed to LLaVA alongside $k$ examples with ground truth labels and a selection of labels to choose from. Then we add 5 steps with further modifications as shown in the top right corner of this figure and according to the additional steps in Table ~\ref{overview}. In Step 1, we add a description of each label as a prefix to LLaVA, the 2nd step is RAG in that we look up the nearest neighbor with CLIP in the KYMKB and retrieve its meme template name, and add this to the text prompt. As a 3rd step, we increase the number of entries in the KYMKB by adding each template example, increasing the coverage of memes we want to look up. We  then filter out retrieved results that are larger than the allowed threshold for each meme template. This is measured by comparing the maximum distance between the base template and examples and the distance between the input meme and the base template. In the fifth and final step, we include the retrieved \textit{about} section in the prompt too.}
    \label{llava_kymkb}
\end{figure*}

\begin{itemize}
    \item Model performance is higher when we exclude the input image but use the extracted OCR on the input meme.
    \item Adding the nearest template title improves performance. 
    \item Adding an explanation of target labels hurts performance.
    \item Excluding information from retrieved KYMKB entities for input memes that are too dissimilar improves performance. Similar to TSplit, dissimilar here means exceeds the threshold value from template to example in the KYMKB. 
    \item Including the \emph{About} section of the nearest neighbor in the KYMKB as input to LLaVA aids in understanding.
\end{itemize} 

\begin{table*}[t]
    \begin{adjustbox}{width=1\textwidth}
    \begin{tabular}{lcccccc}
    \toprule
    \textbf{Method} &  \textbf{MultiOff} & \textbf{Memotion 3 (A)} &\textbf{Memotion 3 (B)} & \textbf{FigMemes} & \textbf{MAMI (A)}& \textbf{MAMI (B)}\\
    \midrule
     
     ICL: Vision and text & 41.6&26.8& 73.8& 25.6 & 48.9& 35.1   \\
     ICL: Vision only & 43 &26.6&72.1& 22.5& 47.6 &33.8   \\
     ICL: text only & 49 &27.3&71.3& 22.8 & 47.7 &34.2  \\
     \midrule
     Using text only \\
     \midrule

     + Label description & 45 &27.7 &72.4 & 23 & 49.4& 33.8   \\
     + Template title & 49 &27.1 &74.1 & 22.3 &51.4  &36.6   \\
     + Template examples & 47 & 29.7 & 73.4 & 23 & 48.6&34.4   \\
     + Filter OOD memes & 53.7& 30.5 & 74.8& 24.7 & 53.7 &36.4   \\
     + \emph{About} section & \textbf{56.4}& \textbf{31.1} & \textbf{75.7} & \textbf{25.4} & \textbf{56.8} &\textbf{37.3}   \\

    \end{tabular}
    \end{adjustbox}
    \caption{Classification results for the best performing version of LLaVA grounded in the KYMKB (\textbf{in bold}) compared against the best-performing method from related work (\textit{in italics}). \emph{+} refers to adding additional input information to the previous row (i.e + template title also includes giving LLaVA the label description in addition to the template title). See remaining tables for full ablations. }
    \label{overview}
\end{table*}

\subparagraph{Parsing LLM output} To calculate the F1 score from the LLM output and ground truth labels, we convert the output and ground truth into binary arrays with $n$ elements with $n$ being the number of classes for the given multilabel task. For multiclass and binary classification tasks, we instead operate with $n-1$ labels. To ensure the model does not start to regurgitate the answer list and input, we restrict decoding to 64 tokens. Similarly, we also ask the model to output the answer "none" if no category is suitable and that "The Assistant must only answer by listing the labels that describe the meme and nothing else.". Finally, if the model does output all the labels, an erroneous answer is given, or no output is recorded, we opt for returning a list of $n$ elements each being 0.

\subparagraph{Examples of LLaVA output} To illustrate the content of the output, we show three examples of LLaVA output on the FigMemes dataset below:

\begin{itemize}
    \item "the meme follows the label set of allusion, exaggeration, and irony. it is not clear which of these labels best fit the ""who killed hannibal?"" meme in this specific image."
    \item "the meme ""distracted boyfriend"" can be classified as an example of anthrop, metaphor, and contrast. in this meme, the man in the image represents the ""distracted boyfriend"" metaphorically, as he appears to be looking at another woman rather than pay"
    \item "the meme follows the label set of allusion, exaggeration, and irony. the labels "i once was blind but now i see " seem to be an allusion to the popular song "i used to be blind""
\end{itemize}

\subparagraph{Discussion}
Table \ref{overview} shows our results. In the case of MultiOff and Memotion 3 (B),  we see that grounding our LLaVA model in the KYMKB improves performance and that the textual modality alone is better than including visual inputs. We believe this is because the offensive or emotional charge signal is stronger in the text. This finding is consistent with \cite{aggarwal-etal-2024-text}, who showed that text alone is enough to detect hate speech in memes. Additional meme-context from the KYMKB can naturally aid in this.
For Memotion 3 (A) and FigMemes on the other hand are more challenging tasks. The model struggles, even with the aid of the KYMKB. This is consistent with our findings for TSplit. In general, LLaVA performs worse than TSplit, thus motivating further study into learning more robust models.

The only exception to the overall trend is MAMI (A) which we believe is due to the LLaMA's strong understanding of misogyny in text. KYMKB is inherently not a knowledge base of misogyny and not a suitable resource for such a message and for the non-templatic memes in MAMI.

\subsubsection{Classification Experiments} 
In this section, we investigate several ablations including how to format the prompt. We examine the following:

\begin{itemize}
\item How do different input modalities impact downstream tasks?
\item Should there be a threshold mechanism for selecting which memes are relevant for meme retrieval?
\item Does including more information from KYM help downstream tasks and do multiple examples help ICL?
\end{itemize}

\subsubsection{Prompt ablations}
Before investigating how LLAVA \cite{liu2023improved} might best perform in prompting experiments on memes, we will describe the overall prompting setup.
To make LLaVA suitable for different tasks on memes without fine-tuning, we perform few-shot in context learning with 3 randomly drawn examples and ground truth answers. These are given to the model when inferring the answer for a given input. The process is illustrated in Table ~\ref{tab:prompt-setup}
However the model struggles with multiple images, even when training and evaluated on the same domain. As such we restrict ourselves to providing the few-shot examples as text input only.

We write a prompt that instructs the model to choose between one or multiple choices from a list of possible answers, which are the labels for a given classification task.

\begin{table}[h]
    \centering
    \adjustbox{width=\columnwidth}{
        \setlength\tabcolsep{2.5pt}
        \newcommand\dottedcircle{\raisebox{-.1em}{\tikz \draw [line cap=round, line width=0.2ex, dash pattern=on 0pt off 0.5ex] (0,0) circle [radius=0.79ex];}}
        \begin{tabular}{l|l}
        \toprule 
        
        {} & \bf Few-shot ICL \\
        \addlinespace[0.1cm]
        Shot &  You are given the following memes with input \\
         &  Optical Character Recognition text: {\color{blue}\texttt{[Text]}} \\
         & Input Image: {\color{blue}\texttt{[Image]}} \\
         & and the following labels {\color{blue}\texttt{[Label]}} \\
         & with explanation {\color{blue}\texttt{[Label Detail]}} \\
         & The meme can be described as {\color{blue}\texttt{[GT Labels]}} \\
          &  \\
         Shot & The next meme has the following input  \\
          & ...  \\
          &  \\
         Input & The final meme has the following input  \\
        & OCR: {\color{blue}\texttt{[Text]}} \\
        Answer & 
        The meme can be described as\textsuperscript{$\star$} {\color{red}\texttt{<answer>}} \\
        
        \midrule 
        \end{tabular}
    }
    \caption{Prompt setup for zero-shot and few-shot prompting with LLaVA} \label{tab:prompt-setup}
\end{table}

\subsubsection{Non-retrieval ablations}
In this section, we focus our attention on whether both image and OCR text or if either one would suffice in helping LLaVA in its downstream classification task. We investigate model performance without retrieving external knowledge by prompting the model with either 1) the OCR text (OR), 2) the original image (IM), or 3) both. We also include explanations of the target labels (LD) to examine if this improves performance.
We report the micro-averaged F1 for each class per task and the average and weighted F1 micro score for all classes. We examine MultiOff, FigMemes, and both MAMI tasks.

\begin{table}[H]
  
  \label{reranking-table1}
  \begin{adjustbox}{width=0.48\textwidth}
  \begin{tabular}{lcccccccc}
    \toprule
    Method & All. & Exa. & Iro. & Ant. & Met. & Con. & F1 & F1(W) \\
    \midrule
OC & 22.7 & 25.2 & 29.9 & 11.5 & 18.2 & 12.4 & 22.8 & 21.6 \\
OC+LD & 26.4 & 24.1 & 31.6 & 14.2 & 5.4 & 16.3 & 23.0 & 20.6 \\ 

\midrule

IM & 20.0 & 29.2 & 27.7 & 13.4 & 17.5 & 16.3 & 22.5 & 21.9 \\ 
IM+LD & 20.2 & 30.7 & 26.2 & 12.9 & 21.8 & 18.0 & 23.2 & 22.9 \\ 

\midrule

IM+OC & 21.1 & 30.8 & 32.7 & 15.6 & 17.4 & 17.1 & 25.6 & 23.8 \\ 
IM+OC+LD & 22.4 & 25.6 & 28.1 & 16.5 & 11.0 & 19.5 & 21.1 & 21.1 \\ 
    \bottomrule
  \end{tabular}
  \end{adjustbox}
  \caption{LLaVA performance in terms of F1 scores on FigMemes. "OC" refers to the text in the meme, "IM" is the meme itself, "ID" refers to our predicted meme template title, "LD" is the label description, i.e a description of what each label means. The different categories are All(usion), Exa(ggeration), Iro(ny), Ant(hropopomorphism), Met(aphor) and Con(trast).}
\end{table}

\begin{table}[H]
  
  \label{mami-table5}
  \centering
    \begin{adjustbox}{width=0.48\textwidth}

  \begin{tabular}{lcccccc}
    \toprule
    Method & sh. & st. & ob. & vi. & F1.& F1(W) \\
    \midrule
OC & 24.3 &  44.4 & 39.6 & 7.0 & 34.2 & 34.1\\
OC+LD & 24.4 & 45.6 & 35.7 & 9.7 & 33.8 & 33.5 \\ 
\midrule
IM & 22.2 & 37.9 & 43.2 & 21.1 & 33.8 & 34.8 \\ 
IM+LD & 23.9 & 42.9 & 42.4 & 18.9 & 34.5 & 36.3 \\ 
\midrule
IM+OC & 22.4 & 41.2 & 43.9 & 15.9 & 35.1 & 35.5\\ 
IM+OC+LD & 27.8 & 42.5 & 46.4 & 19.3 & 37.0 & 38.1\\ 
    \bottomrule
  \end{tabular}
  \end{adjustbox}
  \caption{LLaVA performance in terms of F1 scores on MAMI (B). "OC" refers to the text on the meme, "IM" refers to usage of the image itself, "ID" refers to our predicted meme template title, "LD" refers to label description, i.e a description of what each label means. The categories to predict are Sh(aming), St(ereotype), Ob(jectification) and Vi(olence). We measure the F1 score and the Weighted F1 score, F1(W)}
\end{table}

\begin{table}[H]
  
  \label{multioff-table5}
  \centering
  \resizebox{\linewidth}{!}{ 
  \begin{tabular}{lcccc}
    \toprule
    Method & Of. & N-Of. & F1. & F1(W) \\
    \midrule
OC & 38.7 & 56.3 & 49.0 & 45.6  \\
OC+LD & 19.6 & 58.2 & 45.0 & 34.6  \\ 
\midrule
IM & 17.5 & 56.4 & 43.0 & 36.9 \\ 
IM+LD & 2.1 & 51.2 & 34.9 & 21.2  \\ 
\midrule
IM+OC & 10.3 & 56.7 & 41.6 & 28.4 \\ 
IM+OC+LD & 8.2 & 55.0 & 39.6 & 26.4 \\ 
    \bottomrule
  \end{tabular}
  }
  \caption{LLaVA performance in terms of F1 scores on MultiOff. "OC" refers to the text on the meme, "IM" refers to usage of the image itself, "ID" refers to our predicted meme template title, "LD" refers to label description, i.e a description of what each label means. The two categories to predict are Of(fensive) and Non-offensive (N-Of)}
\end{table}

\subsubsection{Clip-based template retrieval}
As an additional ablation, we test if increasing the size of the CLIP model used for retrieval affects the performance of LLaVA. This model uses a 14x14 patch size on a downscale image of resolution 336x336,

\begin{table}[H]
  
  \label{reranking-table5}
  \centering
    \begin{adjustbox}{width=0.48\textwidth}

  \begin{tabular}{lcccccccc}
    \toprule
    Method & All. & Exa. & Iro. & Ant. & Met. & Con. & F1 & F1(W) \\
    \midrule
OC+ID+LD & 27.3 & 24.5 & 30.2 & 13.6 &5.3 & 12.8 & 22.3 & 20.1 \\  
\midrule
IM+ID+LD & 21.3 & 27.9 & 26.7 & 16.6 & 8.4 & 19.7 & 21.2 & 20.5 \\ 
\midrule
IM+OC+ID+LD & 20.2 & 25.7 & 26.8 & 18.5 & 10.6 & 16.7 & 20.8 & 20.2 \\ 
    \bottomrule
  \end{tabular}
      \end{adjustbox}
      \caption{LLaVA performance in terms of F1 scores on FigMemes. "OC" refers to the text in the meme, "IM" refers to usage of the image itself, "ID" refers to our predicted meme template title, "LD" refers to label description, i.e a description of what each label means.}
\end{table}

\begin{table}[H]
  
  \label{mami-table4}
  \centering
    \begin{adjustbox}{width=0.48\textwidth}

  \begin{tabular}{lcccccc}
    \toprule
    Method & sh. & st. & ob. & vi. & R & F1(W) \\
    \midrule
OC+ID+LD & 24.1 & 47.7 & 42.5 & 13.0 & 36.6 & 37.1 \\ 
\midrule
IM+ID+LD & 23.1 & 42.7 & 45.4 & 20.5 & 35.4 & 37.4 \\ 
\midrule
IM+OC+ID+LD & 22.8 & 38.6 & 46.2 & 22.4 & 34.7 & 36.5 \\ 
    \bottomrule
  \end{tabular}
  \end{adjustbox}
  \caption{LLaVA performance in terms of F1 scores on MAMI (B)}
\end{table}

\begin{table}[h]
  
  \label{multioff-table4}
  \centering
    \resizebox{\linewidth}{!}{ 
  \begin{tabular}{lcccc}
    \toprule
    Method & Of. & N-Of. & F1. & F1(W) \\
    \midrule
OC+ID+LD & 42.4 & 54.2 & 49.0 & 47.1\\ 
\midrule
IM+ID+LD & 15.7 & 56.2 & 42.3 & 31.4\\ 
\midrule
IM+OC+ID+LD & 21.6 & 53.5 & 41.6 & 34.1\\ 
    \bottomrule
  \end{tabular}
  }
  \caption{LLaVA performance in terms of F1 scores on MultiOff}
\end{table}

\subsubsection{Extended clip based template retrieval}
As memes can deviate from their template (see Figure \ref{retrival}, we ask ourselves if including examples of templates from the KYMKB can be used to aid LLaVA in its downstream classification task. To do this, we include both KYMKB templates and example as candidates for retrieval, as the examples may be more similar to the meme in a dataset. Note that we can still access the same information, such as the \textit{about} section, as when using the base template entry because of the KYMKB's structure.

\begin{table}[H]
  
  \label{reranking-table4}
  \centering
    \begin{adjustbox}{width=0.48\textwidth}

  \begin{tabular}{lcccccccc}
    \toprule
    Method & All. & Exa. & Iro. & Ant. & Met. & Con. & F1 & F1(W) \\
    \midrule
OC & 22.7 & 25.2 & 29.9 & 11.5 & 18.2 & 12.4 & 22.8 & 21.6 \\ 
OC+ID & 25.2 & 26.2 & 32.0 & 6.1 & 17.2 & 9.2 & 23.6 & 21.7 \\ 
OC+ID+LD & 26.4 & 24.1 & 31.6 & 14.2 & 5.4 & 16.3 & 23.0 & 20.6 \\ 
\midrule
IM+ID & 20.2 & 30.7 & 26.2 & 12.9 & 21.7 & 18.0 & 23.2 & 22.9 \\ 
IM+ID+LD & 18.1 & 26.0 & 27.4 & 16.8 & 5.0 & 14.4 & 19.5 & 18.5 \\ 
\midrule
IM+OC+ID & 20.1 & 30.4 & 30.2 & 11.3 & 20.5 & 15.6 & 24.3 & 23.0 \\ 
IM+OC+ID+LD & 18.8 & 24.7 & 28.2 & 14.2 & 9.0 & 14.0 & 19.0 & 19.1 \\ 
    \bottomrule
  \end{tabular}
  \end{adjustbox}
  \caption{LLaVA performance in terms of F1 scores for FigMemes. "OC" refers to the text on the meme, "IM" refers to usage of the image itself, "ID" refers to our predicted meme template title,  
  "LD" refers to label description, i.e a description of what each label means.}
\end{table}

\begin{table}[H]

  \label{mami-table3}
  \centering
      \begin{adjustbox}{width=0.48\textwidth}

  \begin{tabular}{lcccccc}
    \toprule
    Method & sh. & st. & ob. & vi. & F1. & F1(W) \\
    \midrule
OC & 22.3 & 46.3 & 38.8 & 5.0 & 33.9 & 33.8 \\ 
OC+ID & 20.3 & 44.3 & 43.6 & 2.6 & 34.3 & 34.1 \\ 
OC+ID+LD & 24.8 & 46.9 & 36.8 & 9.7 & 34.8 & 34.4 \\ 
\midrule
IM+ID & 20.8 & 38.9 & 43.5 & 20.9 & 34.1 & 35.0 \\ 
IM+ID+LD & 24.0 & 41.7 & 43.4 & 20.6 & 34.8 & 36.2 \\ 
\midrule 
IM+OC+ID & 19.7 & 41.9 & 46.6 & 17.4 & 35.9 & 36.5 \\  
IM+OC+ID+LD & 25.6 & 40.2 & 45.0 & 14.8 & 34.9 & 35.8 \\ 
    \bottomrule
  \end{tabular}
  \end{adjustbox}
    \caption{LLaVA performance in terms of F1 scores on the MAMI dataset (Sub-Task B).}
\end{table}

\begin{table}[H]
  \label{multioff-table3}
  \centering
    \resizebox{\linewidth}{!}{ 
  \begin{tabular}{lcccc}
    \toprule
    Method & Of. & N-Of. & F1. & F1(W) \\
    \midrule
OC & 38.7 & 56.3 & 49.0 & 45.6 \\ 
OC+ID & 28.1 & 55.4 & 45.0 & 38.7 \\ 
OC+ID+LD & 34.7 & 55.4 & 47.0 & 42.8 \\ 
\midrule
IM+ID & 13.5 & 53.6 & 39.6 & 29.1 \\ 
IM+ID+LD & 15.7 & 56.1 & 42.3 & 31.4 \\  
\midrule
IM+OC+ID & 10.2 & 56.0 & 40.9 & 28.1 \\ 
IM+OC+ID+LD & 17.3 & 55.7 & 42.3 & 32.3 \\ 
    \bottomrule
  \end{tabular}
  }
  \caption{LLaVA performance in terms of F1 scores on the MultiOff dataset.}
\end{table}

\subsubsection{Clip-based template retrieval filtering}
When we also include the examples of a template from the KYMKB, another question naturally arises: what if there are no suitable entries for a given prompt? To handle such a scenario, we create several filters based on the distance between the base template and its instances. We base this on summary statistics like the interquartile range (IQR), three sigma, mean absolute deviation (MAD), and the maximum distance from template to example, which we find to be the most useful. 
This corresponds to step 4 in \ref{llava_kymkb}. That is, for each template we detect, is this meme in fact within the distribution of this template. 
Note that here we do not use the template examples to detect the template at first, only the templates. We do however use the template examples to make our calculations, which is done using each pairwise distance between the meme template and their examples.

\begin{table}[H]

  \label{reranking-table3}
  \centering
      \begin{adjustbox}{width=0.48\textwidth}
  \begin{tabular}{lcccccccc}
    \toprule
    Method & All. & Exa. & Iro. & Ant. & Met. & Con. & F1 & F1(W) \\
    \midrule
    IQR & & & & & & & &\\
    IM+OC+ID & 18.0 & 30.8 & 31.5 & 16.4 & 18.5 & 14.4 & 24.7 & 22.9 \\  
    \midrule
    Three Sigma & & & & & & & &\\
    IM+OC+ID & 17.8 & 31.2 & 30.8 & 15.1 & 21.3 & 14.6 & 24.8 & 23.2 \\ 
    \midrule
    MAD & & & & & & & &\\
    IM+OC+ID & 20.8 & 31.3 & 30.8 & 14.7 & 20.2 & 16.1 & 25.2 & 23.7 \\ 
    \midrule
    Max & & & & & & & &\\
    IM+OC+ID & 20.5 & 32.0 & 31.4 & 15.5 & 20.2 & 14.2 & 25.4 & 23.8 \\ 
    \bottomrule
  \end{tabular}
  \end{adjustbox}
    \caption{LLaVA performance in terms of F1 scores on FigMemes. "OC" refers to the text in the meme, "IM" refers to using the meme itself, "ID" refers to our retrieved meme template title,  "LD" refers to label description, i.e a description of what each label means.}
\end{table}

\begin{table}[H]
  \label{mami-table2}
  \centering
      \begin{adjustbox}{width=0.48\textwidth}
  \begin{tabular}{lcccccc}
    \toprule
    Method & sh. & st. & ob. & vi. & F1. & F1(W) \\
    \midrule
    IQR & & & & & &\\
    IM+OC+ID & 22.5 & 41.9 & 46.7 & 14.8 & 36.4 & 36.6 \\ 
    \midrule
    Three Sigma & & & & & &\\
    IM+OC+ID & 22.2 & 43.1 & 45.2 & 14.5 & 36.0 & 36.4 \\ 
    \midrule
    MAD & & & & & & \\
    IM+OC+ID & 21.4 & 42.6 & 45.4 & 15.1 & 35.9 & 36.3 \\ 
    \midrule
    Max & & & & & & \\
    IM+OC+ID & 19.7 & 42.8 & 49.0 & 8.1 & 36.4 & 36.3 \\ 

    \bottomrule
  \end{tabular}
  \end{adjustbox}
    \caption{LLaVA performance in terms of F1 scores on the MAMI dataset (Sub-Task B).}
\end{table}

\begin{table}[h]
 
  \label{multioff-table1}
  \centering
    \resizebox{\linewidth}{!}{ 
  \begin{tabular}{lcccc}
    \toprule
    Method & Of. & N-Of. & F1. & F1(W) \\
\midrule
IQR & & & & \\
OC+ID+LD & 43.9 & 60.6 & 53.7 & 50.4 \\
IM+OC+ID & 12.3 & 57.0 & 42.3 & 29.7 \\ 
\midrule
Three Sigma & & & & \\
OC+ID+LD & 46.3 & 56.1 & 51.7 & 50.1 \\
IM+OC+ID & 8.3 & 56.4 & 40.9 & 27.1 \\ 
\midrule
MAD & & & & \\
OC+ID+LD & 42.4 & 54.2 & 49.0 & 47.1 \\
IM+OC+ID & 12.0 & 55.6 & 40.9 & 29.0 \\ 
\midrule
Max & & & & \\
OC+ID+LD & 35.4 & 50.0 & 43.6 & 41.1 \\
IM+OC+ID & 10.1 & 55.3 & 40.2 & 32.7 \\ 

    \bottomrule
  \end{tabular}}
   \caption{LLaVA performance in terms of F1 scores on the MultiOff dataset.}
\end{table}

\newpage

\subsubsection{Extended meme information}
We choose the maximum distance algorithm as the default method of selecting relevant meme content based on its simplicity and performance in filtering experiments above. We now investigate including the \emph{about} section as additional information to add to our prompt and if this grounds the LLM in meme knowledge.

\begin{table}[H]
 
  \label{reranking-table2}
  \centering
      \begin{adjustbox}{width=0.48\textwidth}

  \begin{tabular}{lcccccccc}
    \toprule
    Method & All. & Exa. & Iro. & Ant. & Met. & Con. & F1 & F1(W) \\
    \midrule 
    IQR & & & & & & & &\\
    \midrule
OC+ID+KYM  & 24.7 & 28.2 & 33.3 & 14.3 & 20.0 & 16.5 & 25.4 & 24.4 \\ 
\midrule
    Max & & & & & & & &\\
IM+OC+ID+KYM & 20.3 & 30.4 & 30.1 & 12.2 & 20.6 & 19.1 & 24.9 & 23.6 \\ 

    \bottomrule
  \end{tabular}
  \end{adjustbox}
   \caption{LLaVA performance in terms of F1 scores on FigMemes. "OC" refers to the text in the meme, "IM" refers to usage of the meme itself, "ID" refers to our retrieved template title, "LD" is the label description, i.e a description of what each label means.}
\end{table}

\begin{table}[H]
  
  \label{mami-table1}
  \centering
      \begin{adjustbox}{width=0.48\textwidth}

  \begin{tabular}{lcccccc}
    \toprule
    Method & sh. & st. & ob. & vi. & F1. & F1(W) \\
    \midrule
    IQR & & & & & & \\
OC+ID+KYM  & 24.9 & 45.3 & 47.2 & 8.3 & 37.8 & 37.3 \\ 
IM+OC+ID+KYM & 21.5 & 41.0 & 43.7 & 14.3 & 34.5 & 35.0 \\ 

    \bottomrule
  \end{tabular}
  \end{adjustbox}
  \caption{LLaVA performance in terms of F1 score on the MAMI dataset (Sub-Task B).}
\end{table}

\begin{table}[H]
 
  \label{multioff-table2}
  \centering
  \begin{tabular}{lcccc}
    \toprule
    Method & Of. & N-Of. & F1.  & F1(W) \\
    \midrule
IQR & & & & \\
OC+ID+KYM  & 55.2 & 57.5 & 56.4 & 56.1 \\ 

    \bottomrule
  \end{tabular}
   \caption{LLaVA performance in terms of F1 scores on the MultiOff dataset (Sub-Task B).}
\end{table}

\end{document}